%% file: main.tex
\documentclass{article} 
\usepackage[preprint]{neurips_2023}
\newif\ifarxivversion
\arxivversiontrue

\ifarxivversion
\newcommand\allthenoiseurl{\url{https://all-the-noises.github.io}}
\else
\newcommand\allthenoiseurl{\url{https://all-the-noises-anony.github.io}}
\fi

\usepackage[utf8]{inputenc} 
\usepackage[T1]{fontenc}    
\usepackage[hidelinks]{hyperref}       
\usepackage{url}            
\usepackage{booktabs}       
\usepackage{amsfonts}       
\usepackage{mathtools}
\mathtoolsset{showonlyrefs,showmanualtags}
\usepackage{fancyvrb}
\fvset{frame=single, framesep=3mm, fontsize=\small}
\usepackage{amsthm}
\newtheoremstyle{examplestyle}
  {6pt plus 2pt minus 2pt}  
  {6pt plus 2pt minus 2pt}  
  {\normalfont}  
  {}      
  {\bfseries} 
  {.}     
  {.5em}  
  {}      
\theoremstyle{examplestyle}

\usepackage{nicefrac}       
\usepackage{microtype}      
\usepackage{xcolor}         

\usepackage{hyperref}
\usepackage{ifthen}

\usepackage{natbib}
\usepackage{graphicx}
\usepackage{xspace}

\usepackage{amsmath,amsthm,amsfonts,amssymb}

\usepackage{algorithm}
\usepackage{algpseudocode}
\usepackage{enumitem}
\usepackage{parskip}

\input{std-macros.tex}

\input macros.tex

\renewcommand{\cite}{\citep}

\title{Measuring all the noises of LLM Evals }
\author{
  Sida I. Wang \\
  \texttt{sida@meta.com} \\
  FAIR at Meta
}

\begin{document}

\maketitle
\begin{abstract}
   Applying well-established statistical methods effectively to LLM evals requires consideration of model correlations and prediction noises. Leading evals often ignore data noise, have low statistical power, or make unnecessary trade-offs. We  define and measure three types of noise: 1) prediction noise from generating different answers on a given question, 2) data noise from sampling questions, and 3) their combined total noise following the law of total variance.
   To emphasize relative comparisons and gain statistical power, we propose \emph{all-pairs comparison}, which applies paired comparison to all pairs of LLMs and measures all the noise components based on millions of question-level predictions across many evals and settings, revealing clear trends. First, most evals exhibit a characteristic and highly predictable total noise level across all model pairs. Second, paired prediction noise typically exceeds paired data noise, which means reducing prediction noise by averaging can significantly increase statistical power. By measuring all the noises together, we can assess eval results in context and make better empirical decisions. 
\end{abstract}

\section{Introduction}
Identifying signal from noise is important for interpreting any experiment results. 
Statistical methods can increase the rigor of the conclusions drawn from data \cite{benjamini2021asa}.
Although the methods are well-established \cite{wasserman2004all, miller2024error} and there is great interest in measuring noise, leading LLM evals still ignore the data sampling noise \cite{merrill2026terminalbench} (Example~\ref{ex:direct_prediction_noise}), have low statistical power \cite{dubey2024llama3} (Example~\ref{ex:humaneval}), or make unnecessary trade-offs \cite{chiang2024chatbot, gu2024cruxeval} (Example~\ref{ex:confidence}). By using the global identity of the models and LLM eval questions to measure all types of noises together, we increase their rigor, statistical power and provide a more complete picture of noise.%
\footnote{Interactive results based on 100k+ model pairs: \allthenoiseurl}


To measure noise without confusion, we must understand the prediction noise of LLMs and that it is not the only noise. Unlike previous probabilistic models that typically have a clear best prediction, LLMs produce persistently diverse predictions.
For a model $A$ evaluated on questions $x$ with prediction seed $\seed$, we have
\begin{align}
\quad \var_{x,\seed}[A(x, \seed)] &= \var_x[\E_\seed[A]] + \E_x[\var_\seed[A]],\\
\text{Total variance} &= \text{data variance} + \text{prediction variance}.
\end{align}
The \emph{prediction variance} $\E_x[\var_\seed[A]]$ is due to the model generating different answers on a fixed question. The \emph{data variance} $\var_x[\E_\seed[A]]$ measures how the questions vary in difficulty level and thus vary in their expected metrics. Prediction variance can be reduced but, in practices, usually persists. Data variance reflects the reliability beyond the specific set of questions.
While prediction variance can be reduced by averaging, data variance is irreducible and can be large as the questions vary in difficulty.





While questions range in their difficulty, similar models tend to do similarly on each question, so a greatly reduced noise is possible only if \emph{paired comparison} is used. The combination of paired comparison and prediction noise reduction reduce the noise by 3-6 times and 4 times in Example~\ref{ex:humaneval} and Example~\ref{ex:trainingcurve}b (training curve).
The difficulty is that paired comparison only applies to pairs of models, rather than per model, and thus cannot be directly adopted on leaderboards.

In this work, we propose the all-pairs comparison to measure all the types noise, which uses paired comparisons on all pairs of models to measure the data, prediction, and total paired variances $\Var[A-B]$, for all model pairs $A, B$. Across many evals and more than hundred thousand model pairs in different settings, our method reveals clear patterns in the noise measurements. 

\begin{figure}[t]
    \centering
    \includegraphics[scale=0.13, trim={0pt 50pt 322pt 40pt}, clip]{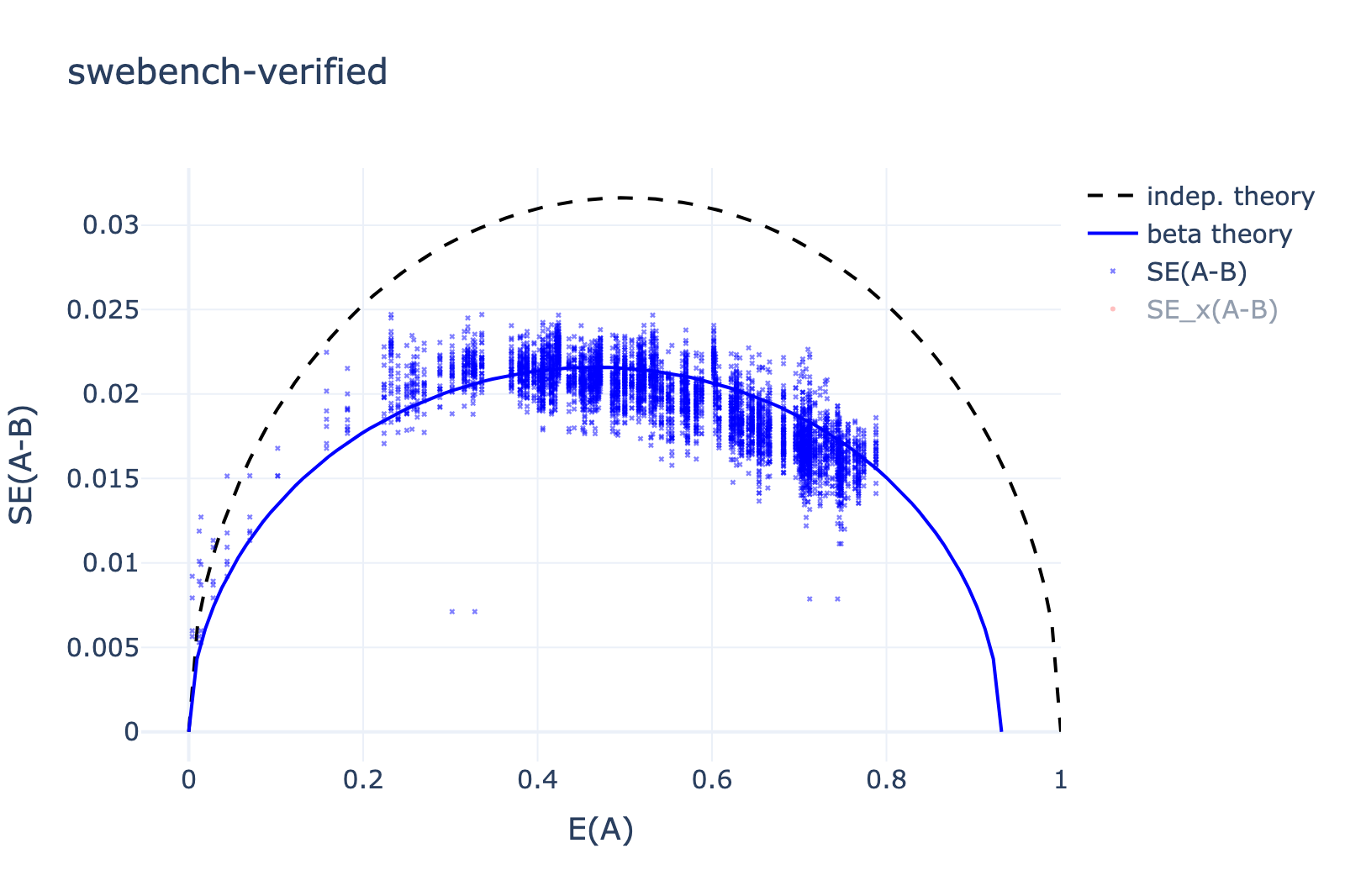}
    \includegraphics[scale=0.13, trim={0pt 50pt 26pt 40pt}, clip]{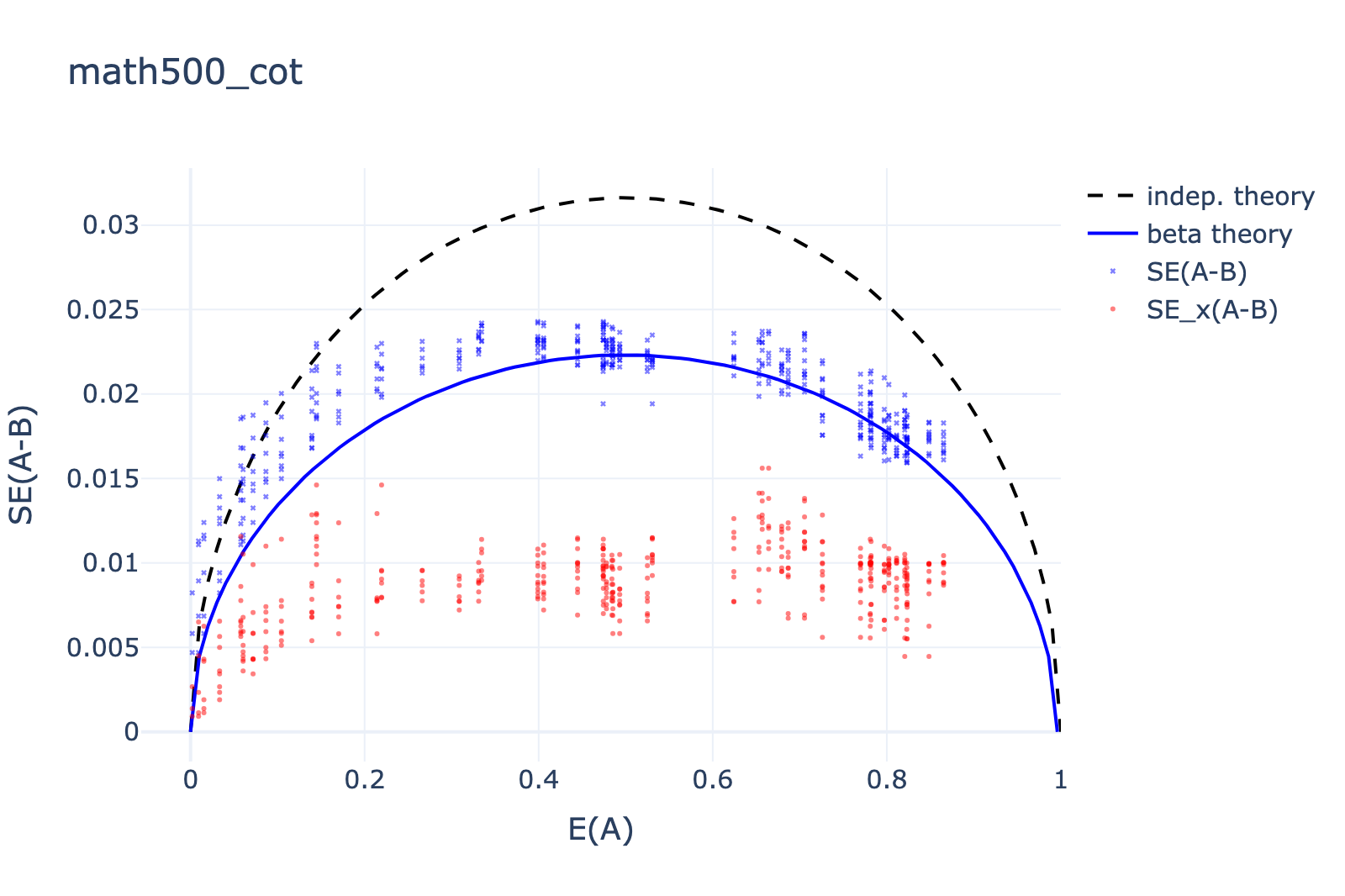}

    \caption{paired total standard errors \textsf{SE(A-B)} vs. the accuracy \textsf{E(A)}. Each dot is a paired comparison of model A vs. B, agreeing with the beta theory (trend line). Left: on SWEBench estimated from 1 sample per question, Right: on MATH500 from multiple samples per question and thus also showing the irreducible data noise \textsf{SE\textunderscore{}x(A-B)}.
    }
    \label{fig:pairedSE}
\end{figure}

For a simple total noise analysis, we recommend Figure~\ref{fig:pairedSE} Left, showing the paired total noise of all model pairs on SWEBench-Verified against the beta theory prediction.
We see that most evals have a characteristic total noise level that holds across all pairs of LLMs or agents. 
Specifically, the paired test usually gives a total paired variance similar to the basic variance predicted by the accuracy $\E[A]$ of model $A$, resulting in this rule of thumb for many evals
\begin{align}
    \Var[A - B] &\approx \Var[A] = \E[A](1-\E[A]), \\
    \SE[A - B] &\approx \SE[A] = N^{-1/2}\Var[A]^{1/2}.
\label{eq:ruleofthumb}
\end{align}
When a trend holds, we have accurate-enough error bars and we can even infer the likely significance of results reported by others in the context of these paired comparisons. 

Next, we measure the paired data noise, prediction noise, and total noise. Figure~\ref{fig:pairedSE} Right shows that the data noise on MATH500 is significantly smaller than the total noise, which means that reducing the prediction noise can lead to a significant increase in statistical power, or a significant decrease in the minimum detectable effect size (Example~\ref{ex:humaneval}).
This turns out to be typical in other evals and settings, we find that the paired prediction noise is typically greater than the paired data noise. Even greater reductions can be expected for controlled experiments between similar model pairs. When the prediction noise is greater, inferring the noise from the training curve becomes useful (Example \ref{ex:trainingcurve}).
The importance of these choices means that we must specify what noise we are measuring: data, prediction, or total noise using paired vs. unpaired comparison; otherwise we remain confused and cannot interpret reported noise measurements \cite{chan2025mlebenchevaluatingmachinelearning, merrill2026terminalbench}. 

We speculate that the uniformity of these results comes from the relative similarity among LLMs training on similar data using similar methods. Like models, eval questions also have a global identity analogous to the ability to repeat experiments on the same subjects anywhere. This allows the all-pairs comparison to be both informative and generalizable.
The results are more informative than error bar per model, where any exceptions are revealed (\S\ref{sub:allpair}, \S\ref{sub:exceptions}) and all results can be explored interactively. 

\paragraph{Contributions.}
We clearly define and empirically measure prediction noise, data noise, and total noise on hundred thousands of model pairs. We propose all-pairs comparisons to increase the rigor, statistical power and provide a more complete picture of all the noises in evals.
Our general findings from these tools allow practitioners to assess results in a larger context, often without custom statistical testing.

\ifarxivversion
\paragraph{Contents.} We present the main concepts informally and through examples in \S\ref{sec:key}; then more precisely in \S\ref{sec:method}. 
\S\ref{sec:experiment} describes exploratory data visualization methods based on all-pairs comparison.
For discussions, \S\ref{sub:recs} provide recommendations for practitioners to better deal with noise; \S\ref{sub:hardproblems} on why harder questions cannot yet compensate for small sample sizes; \S\ref{sub:standard} for some caveats of averaging; \S\ref{sub:mainremarks} explains why we observe these results. To ensure correctness, we validate our method by comparing with the well-established bootstrap and sign test methods (\S\ref{sec:equivalence}), and test our sample estimators (\S\ref{sec:estimator_testing}) against several generative models as well as on real LLM evals.

\fi



\input{related_work.tex}
\input{methods.tex}

\input{validation}
\input{exploratory}

\input{discussions}

\bibliography{main}
\bibliographystyle{acl_natbib}
\clearpage
\appendix
\input{A-method}

\input{A-beta}
\input{A-caveats}

\input{A-recs}

\input{A-equivalence}
\input{A-testing}

\end{document}

%% file: std-macros.tex
\newcommand{\var}{\mathrm{Var}} 
\newcommand{\cov}{\mathrm{Cov}} 

\newcommand\R{\ensuremath{\mathbb{R}}} 
\newcommand\eqdef{\ensuremath{:=}} 
\newcommand{\1}{\mathbb{I}} 

\ifthenelse{\isundefined{\definition}}{\newtheorem{definition}{Definition}}{}
\ifthenelse{\isundefined{\assumption}}{\newtheorem{assumption}{Assumption}}{}
\ifthenelse{\isundefined{\hypothesis}}{\newtheorem{hypothesis}{Hypothesis}}{}
\ifthenelse{\isundefined{\proposition}}{\newtheorem{proposition}{Proposition}}{}
\ifthenelse{\isundefined{\theorem}}{\newtheorem{theorem}{Theorem}}{}
\ifthenelse{\isundefined{\lemma}}{\newtheorem{lemma}{Lemma}}{}
\ifthenelse{\isundefined{\corollary}}{\newtheorem{corollary}{Corollary}}{}
\ifthenelse{\isundefined{\alg}}{\newtheorem{alg}{Algorithm}}{}
\ifthenelse{\isundefined{\example}}{\newtheorem{example}{Example}}{}

\newcommand{\E}{\ensuremath{\mathbb{E}}} 

%% file: macros.tex
\newcommand{\Var}{\mathrm{Var}}
\renewcommand{\E}{\mathrm{E}}

\newcommand{\SE}{\operatorname{SE}}
\newcommand{\Beta}{\operatorname{Beta}}
\newcommand{\pred}{\text{pred}}

\renewcommand\xi{\ensuremath{\epsilon}}
\newcommand\seed{\ensuremath{\xi}}

%% file: related_work.tex
\section{Related work}

There is a strong interest to put error bars on leaderboard and results, Chatbot arena \cite{chiang2024chatbot} and CRUXEval \cite{gu2024cruxeval} computed confidence intervals using bootstrap
which allowed them to obtain per-model confidence intervals with trade-offs (Example~\ref{ex:confidence}). For example, Chatbot arena aggregates comparisons with all other models to compute their confidence intervals, which necessarily loses any special properties of a particular pair of models. \citet{gu2024cruxeval} uses a fixed reference model which also fails to capture any special pairs.
Llama 3 \cite{dubey2024llama3} uses unpaired Bernoulli confidence intervals, which is much looser than paired comparison (Example~\ref{ex:humaneval}) and does not separate the data noise from the prediction noise.
Works measuring the noise across runs \cite{merrill2026terminalbench} or random seed \cite{madaan2024variance} on the same eval necessarily ignores the critical data noise (Example~\ref{ex:trainingcurve}).
This work compares all pairs of models, and uses the summary statistics of all the comparisons to get meaningful noise measurements. More importantly, we show that it is possible to get meaningful graphical summaries and do this in a separable way (\S\ref{sub:recs}).

\citet{hagele2026hotmess} propose a bias-variance decomposition on KL, and studied the ratio of prediction noise to total noise. Recent works \cite[][among others]{romanou2026brittlebench, haase2026withinmodel, messing2026hidden} proposed more detailed decompositions and study sources of variations such as prompt and judge. \citet{miller2024error} considered a impactful clustered correction. This work attempts to keep the core decomposition method as simple as possible to focus on what can be learned from the empirical all-pairs results and correctness (\S\ref{sec:A.method},\ref{sec:estimator_testing}).

Much work analyzed noise in science and machine learning \cite[][among others]{lehmann2005testing, dror-etal-2018-hitchhikers, hermann2024experimental, card-etal-2020-little}. We adopt the approach of \citet{miller2024error} to estimate the variance directly. While using samples and paired comparisons is standard statistics, their approach is clarifying.
Paired comparison is described by \citet{wasserman2004all} in Wald test examples.
This method is also consistent with the bootstrap \cite{efron1979bootstrap} and the sign test \cite{dixon1946sign} when applicable (\S\ref{sec:equivalence}). \citet{efron1986bootstrap} clearly described estimating the standard error as the example of not needing bootstrap. The concept of total variance decomposition is present in the \emph{analysis of variance}, but Section 3 of \citet{miller2024error} applied this concept to LLMs under \emph{variance of
the conditional mean} and \emph{mean conditional variance}, which we call the data noise and prediction noise. They also advocated averaging (resample), but do not provide any data on how much this matters. In this work, we develop and test the methods to directly estimate the noise types from samples and measure these components on actual evals and models. 




%% file: methods.tex
\section{Main concepts and examples}
\label{sec:key}


\subsection{Types of noise}
We should clearly distinguish three types of noise: the prediction sampling noise when answering a given question, the data sampling noise due to using a finite sample from the population of possible questions, and their sum total noise.
\paragraph{Prediction noise.}
The prediction sampling noise comes from generating different answers on a given question with the same model. LLMs have the remarkable ability to sample independent predictions that are especially interesting on reasoning and agentic questions. 
In the physical world, the prediction noise typically cannot be measured directly --- humans cannot completely forget their previous thoughts and medical treatments cannot be undone to get more independent samples. Previous probabilistic ML models such as classifiers can typically make their best prediction directly and do not have interesting prediction noise.

The prediction noise is also persistent even though we can try to reduce it by fixing the random seed, decreasing the temperature, averaging over many samples per question, or applying an aggregation method such as majority voting and verification.
Fixing the random seed or lowering the temperature is brittle, which only works on the exact same model. Trying to produce the best answer by aggregation is perhaps the most sound method, though this still leaves some prediction noise due to the diversity of samples.
Among these noise reductions, only averaging and fixing the random seed gives the same mean while others change the mean as well. 

Other than sampling, additional prediction noise can come from inference-time choices such as prompt formatting, hyperparameters, and answer extraction; and training-time choices such as the random seeds, hyperparameters, and training steps. All of these can in theory be measured directly by changing the seed (for each question) then drawing a new prediction.  While training is not typically considered a noise, it can result in more noise than signal over a small number of steps.

\begin{example}[Measuring prediction noise directly]
\label{ex:direct_prediction_noise}
Run the model $K$ times to get average scores $\bar{A}_1, \ldots, \bar{A}_K$ is the average score on the same set of  questions on the $k$-th run of the eval. We then take $\operatorname{std}(\bar{A}_1, \ldots, \bar{A}_K)$. This is sometimes used as the only measure of noise \cite{merrill2026terminalbench, madaan2024variance}. \S\ref{sub:alternative_decompose} show that this can be quite inaccurate and how we can substantially improve this estimator using question level information.
\end{example}

If we only care about a specific set of questions, then we only need to measure the prediction noise. This is not appropriate unless the goal is to memorize a specific set of questions (e.g. driving knowledge test where the set of questions is published), or to solve specific important open problems (AGI/ASI, P vs. NP). On most evals, we care about the underlying ability estimated by the eval rather than the specific questions in the eval, so the data noise must be accounted for.

\paragraph{Data noise.}
The eval data sampling noise comes from sampling a particular set of $N$ questions instead of another equally good $N$ questions from the unobservable population of questions. As long as the model has different expected accuracy on different questions (i.e. some questions are harder), the mean will vary when a different set of questions is sampled. 
Unlike the prediction noise, the data noise cannot be directly measured by evaluating on the fixed, given set of questions; and it cannot be reduced. Though the data noise can be inferred using the bootstrap or the variance estimator, it is not observed in experiments such as the training curve (Example~\ref{ex:trainingcurve}) or empirically varying the training seeds \citep{madaan2024variance}. 

To see the presence of data noise, suppose that out of 100 True and False questions, model $A$ is correct on 50 random questions, $B$ is correct on 52 random questions out of the same 100 questions. Then $B$ is not reliably better than $A$ even when each question is answered deterministically with no prediction noise. Suppose we have $p=0.5$ chance of sampling a question that is answered correctly by $A$, then the number of correctly answered questions has a standard deviation of $(100 p(1-p))^{1/2}=5$, and thus it is typical to see a difference of more than 2 questions.
Statistical methods such as computing the standard error or bootstrap focus on the data noise and will show that $A$ is not reliably better in this case.

\paragraph{Total noise = prediction + data noise.}
We can make this relation precise using the law of total variance. The extensive literature about noise focuses on the total noise or the data noise, perhaps because the prediction noise is uninteresting as in classification, or not measurable directly as in most non-digital experiments.

\paragraph{Paired comparison.}
Whereas insights, skills, and resources are required to improve LLM ability, paired comparison can increase the signal significantly. Thus they should be recommended, especially given the enormous resources devoted to improving models. Paired comparison makes use of prediction differences on the same question to gain more information.
To see why paired comparison is better on real data, Figure~\ref{fig:heatmap} shows the heatmap of correctness where models at a similar overall accuracy (close in $x$-axis) also do similarly on the individual questions. Thus, paired comparison has the potential to reduce the data noise significantly.
Examples~\ref{ex:trainingcurve}b and \ref{ex:humaneval} show the potential increase in power using paired comparison, especially when prediction noise > data noise, leading to the ability to detect differences several times smaller.

\paragraph{Paired vs unpaired intuition.}
In the unpaired case, suppose $A$ scored $50\%$ on this year's exam, $B$ scored $60\%$ on last year's exam where each exam contains different questions drawn from the same distribution. So $B$ might have done better because last year's exam contained easier questions. In the paired case, suppose $B$ scored $60\%$ on the same exam as $A$ containing exactly the same questions, then the same $10\%$ difference is more significant. 

\begin{example}[HumanEval]
\label{ex:humaneval}
HumanEval \cite{chen2021evaluating} has $N=164$ questions. If a model has $\bar{A} = 0.5$ accuracy, then the unpaired standard error (SE) is $\sqrt{ 1/164 \times 0.5 (1-0.5)}=$ 4\%. Since a factor of $\sqrt{2}$ is needed when comparing pairs of models, we need $ 4\% \times \sqrt{2} \times 1.96 = 11\%$ to get $0.05$ $p$-value. The table below shows the reductions with paired and averaging, and how the combination can significantly reduce the minimum detectable difference between typical pairs of LLMs.
\begin{table}[h]
\centering
\begin{tabular}{lrr}
Test Type & $\SE(\bar{A}-\bar{B})$ & Difference for $p<0.05$ \\
\midrule
Unpaired & 6\% & 11\% \\
Unpaired + averaging & 4\% & 8\% \\
Paired  & 4\% & 8\% \\
Paired + averaging & 1--2\% & 2-4\% \\
\end{tabular}
\end{table}

Table 10 of Llama 3 \cite{dubey2024llama3} reports the unpaired 95\% confidence intervals of around $7\%$, thus many of the comparisons do not clear the $\sqrt{2}\times 7\%$ difference with their method. 
\end{example}

\begin{example}
\label{ex:trainingcurve}
\textbf{a) The training curve is invalid in theory. }
The \emph{training curve} plots the eval result as a function of training steps and graphically shows the variance of eval results as a function of training steps (Figure~\ref{fig:traincurve}, col 1). While separated training curves are intuitively needed for reliable results, it is not sufficient. The training curve is insufficient because it only shows the prediction noise and completely ignores the data noise as it evaluates on the same questions repeatedly. In theory, the data noise can be much higher than the prediction noise. The prediction noise can be zero with deterministic predictions and convergent training. 

\textbf{b) The training curve is valid when prediction noise is dominant.}
In practice, if the prediction noise dominates the data noise, then the training curve method is valid. Figure~\ref{fig:traincurve} shows increasingly powerful analysis on the training curve \ifarxivversion\cite{wei2025trainingsuperintelligentsoftwareagents}\fi{} visualized by the bootstrap.

\begin{figure}[thb]
    \centering
    \includegraphics[width=1\textwidth, trim={20pt 55pt 30pt 155pt}, clip]{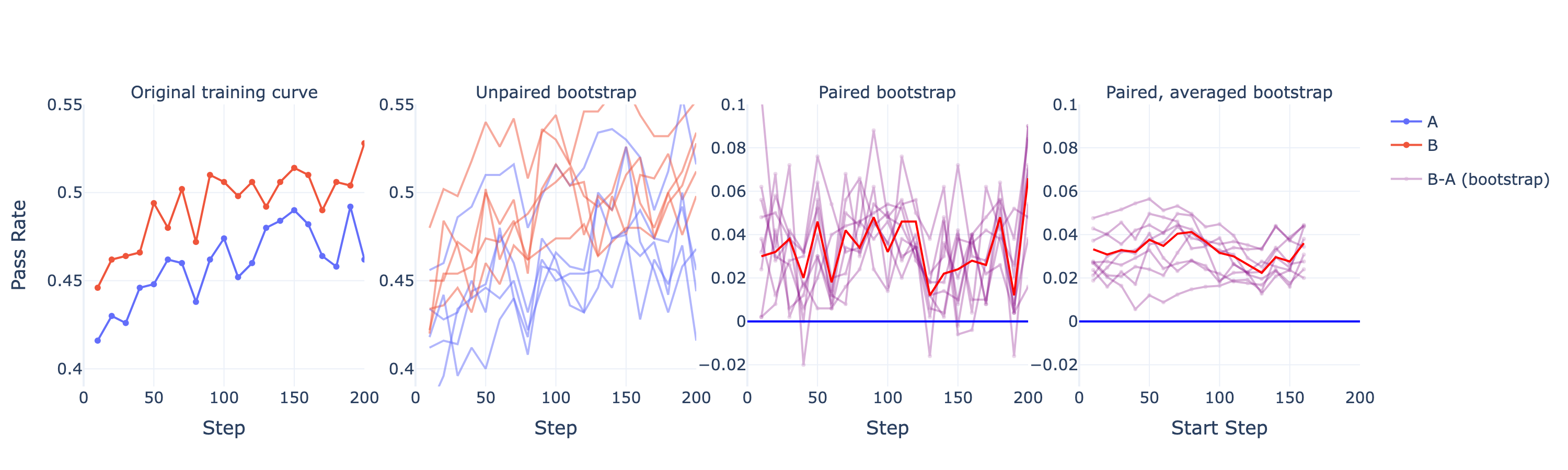}
    \caption{Increasingly sensitive training curves on SWEBench-Verified with $N=500$, A and B both start from the same checkpoint at step 0. \textbf{Column 1) Original training curve.} One prediction accuracy, showing B is better than A over all steps. \textbf{2) Unpaired bootstrap.} A is better than B on many resampled set of 500 questions, showing a meaningless difference using this analysis method.
    \textbf{3) Paired bootstrap.} Slightly significant with a $z$-score of 1.7 at 1 prediction per question. \textbf{4) Paired, averaged bootstrap.} The result is very significant with a $z$-score of 3.5 when the question level predictions are averaged over 5 consecutive training checkpoints to reduce the prediction noise.}
    \label{fig:traincurve}
\end{figure}
\end{example}

\paragraph{All-pairs comparison.}
We measure the paired standard error $\SE(A-B)$ of all model pairs $A$ and $B$, where the total noise is always measured, and the data and prediction noises are measured whenever possible.
While this approach can be inconclusive in general (Example~\ref{ex:confidence}), our measurements show that on LLM evals this turned out to be quite informative, so we highlight the following two findings.

\paragraph{Finding: predictable total noise.}
Figure~\ref{fig:pairedSE} shows the total noise vs. the accuracy of $A$. We observe that the total noise is quite predictable and agrees well with a theoretical model. This means each eval has a characteristic total noise as a function of the accuracy. This is the right noise number to use for experiments where one prediction is made per question.

\paragraph{Finding: typically prediction noise $>$ data noise.} When this holds, we can gain much statistical power by using the expected metric where the prediction noise is averaged away to be comparable to the data noise remaining. Figure~\ref{fig:pairedSE} Right shows that on MATH500, the data standard error (SE) is typically less than $1/2$ of the prediction SE between different final models. Example~\ref{ex:trainingcurve}b and Figure~\ref{fig:traincurve} show training curves evaluated on SWEBench-verified starting with the same checkpoint, the data SE is $\sim1/6$ of the prediction SE. If we average over 36 predictions, then the total SE would be $\sqrt2 / 6 = 0.24$, allowing us to detect effects that are $0.24$ the size with the same $p$-value and power. Example~\ref{ex:humaneval} shows that the reduction on HumanEval is smaller but still a factor of 2. 
For controlled experiments, this allows us to detect differences several times smaller at the same significance level and power. Important caveats are discussed in \S\ref{sub:standard}.


\begin{example}[Error bars for leaderboards]
\label{ex:confidence}
\citet{chiang2024chatbot, gu2024cruxeval} made trade-offs to put error bars on a leaderboard. To do this, they have to aggregate all paired comparisons into one number by using ELO or a reference model, which is necessarily lossy. This can be incomplete: let $A$ be the baseline, $B$ and $B'$ are a pair of similar predictions very different from $A$, so $\var[A-B] \gg \var[B-B']$. ELO similarly merges head-to-head comparisons ($B$ vs $B'$) with background comparisons, while still leaving prediction noise unspecified. This example shows it's not in general possible to have one error bar per model on a list for comparisons. Using the all paired comparisons like in Figure~\ref{fig:pairedSE}, we can check if meaningful error bars per model is possible, identify any surprises, and see any statistical claims in context.
\end{example}


\section{Method}
\label{sec:method}
\subsection{Setup and notations}
To make these concepts precise, suppose there are $N$ eval questions $\{x_1, \ldots, x_N\}$, which are typically text prompts.
The model makes prediction $\hat{y}(x)$ and we get $\operatorname{metric}(\hat{y}, x)$, typically 0 for incorrect and 1 for correct but can be a real number or an aggregate metric. Any information needed to compute the metric such as the ground truth answer or agent environments are included implicitly.
We use $A(x) \eqdef \operatorname{metric}(\hat{y}(x), x)$ to denote the metric function evaluating model $A$ on the question $x$. 


On an eval, we compute the average of all $N$ questions to get the mean metric
\begin{align}
\bar{A} \eqdef \frac{1}{N} \sum_{i=1}^N A(x_i) \approx \E_x[A(x)].
\end{align}

The estimated variance of $A$ and standard error (SE) of the mean are respectively
\begin{align}
\var_x[A(x)] &\approx \frac{1}{N} \sum_{i=1}^N \left(A(x_i) - \bar{A} \right)^2, \\
\SE(A, N) &= N^{-1/2} \cdot \var[A]^{1/2},
\end{align}
where the standard error is normalized to the right scale for comparison to the effect size. We just say \emph{noise} when there is no need to distinguish the variance and the standard error.
We avoid the convention that corrects the bias and increase the error of $\var_x[A]$ using $1/(N-1)$. 

With models $A$ and $B$, the score difference $\bar{A} - \bar{B}$
can be compared to $\SE(A-B) = (\SE(A)^2 + \SE(B)^2)^{1/2}$ to determine if the difference is likely due to chance. While this is simple and correct, it is also unnecessarily weak (Example~\ref{ex:humaneval}). 

\paragraph{Paired.}
Whereas innovation, insight, or compute is required to get real signal, tightening the analysis is easy with paired comparison and should be recommended generally.
$A(x)$ and $B(x)$ are correlated when the same question is used to evaluate both $A(x)$ and $B(x)$. 
To be more powerful for free, we can use the paired variance
\begin{align}
\var_x[A(x)-B(x)] = \var_x[A] + \var_x[B] - 2\cov_x[A(x), B(x)].
\end{align}
The $\cov$ term may potentially reduce the paired variance to 0 when $A$ and $B$ are perfectly correlated.

From the standard errors we can obtain the $z$-score $z = \frac{\bar{A} - \bar{B}}{\SE[A-B]}$. When $N > 100$ is moderately large, then the Central Limit Theorem applies and the $z$-scores can be converted to confidence intervals and $p$-values. For example, $\Pr[|z| > 1] = 0.32$ is very weak, $\Pr[|z| > 5] < 10^{-6}$ is beyond doubt, and $\Pr[|z| > 1.96] = 0.05$ for a $p$-value of $0.05$ is a conventional standard.



\subsection{Sampling multiple predictions on each question}

To capture this setting, we use $\seed$ for the random seed that generates the prediction, which is independent of the question $x$, thus allowing the concise notation $\var_x[\E_\seed[A]] = \var_{x} [E_{\seed|x}[A(x, \seed) \mid x]]$.
As before, but now also averaging over the $K$ samples for each question $x$, we consider the average score
\begin{align}
\bar{A}^{(K)} &= \frac{1}{N}\sum_{i=1}^N  \frac1{K} \sum_{j=1}^{K} A(x_i, \seed_{ij}) \approx \E_{x, \seed}[A(x, \seed)],
\end{align}
The estimated variance is then
\begin{align}
\var_{x, \seed}[A(x, \seed)] &\approx \frac{1}{N}\sum_{i=1}^N  \frac1{K} \sum_{j=1}^{K} (A(x_i, \seed_{ij}) - \bar{A}^{(K)})^2.
\end{align}
One choice is to consider the $K$-run average $\bar{A}^{(K)}$ as the metric function \cite{miller2024error}. We prefer to remove the dependency on $K$ by decomposing into components,
\begin{align}
\var_{x, \seed}[A] = \var_x[\E_\seed[A]] + \E_x[\var_\seed[A]].
\end{align}

$\var_{x, \seed}[A]$ is the total variance of first drawing a question $x$ from the pool of questions, and then drawing a sample answer for this question. $\var_x[\E_\seed[A]]$ is the variance of the expected score $\E_\seed[A]$, which we call the data variance.
$\E_x[\var_\seed[A]]$ is the variance due to sampling from the model, which we call the prediction variance. See \S\ref{sub:alternative_decompose} for why $\E_x[\var_\seed[A]]$ is the same as the prediction variance that can be directly measured on a fixed eval. 

All the variances can be normalized to be their respective standard errors,
\begin{align}
\SE(A, N) &= N^{-1/2} (\var_{x, \seed}[A] )^{1/2}, \\
\SE_x(A, N) &= N^{-1/2} (\var_x[\E_\seed[A]])^{1/2}, \\
\SE_\pred(A, N) &= N^{-1/2} (\E_x[\var_\seed[A]])^{1/2}.
\end{align}

\paragraph{Paired.}
When we have two models $A$ and $B$, we can consider the paired variance $\var_{x, \seed}[A(x, \seed) - B(x, \seed)]$,
which also satisfies the law of total variance on $A-B$,
\begin{align}
\var_{x, \seed}[A - B] = \var_{x}[\E_\seed[A - B]] 
+ \E_{x}[\var_\seed[A - B]].
\label{eq:paired_total}
\end{align}
See~\S\ref{sec:A.method} for how to estimate from samples, the importance of small-$K$ correction, and implementation details.


%% file: exploratory.tex
\section{Experiments}
\label{sec:experiment}
The full experimental results of this paper can be found at \allthenoiseurl{} for higher reproducibility and freshness than is possible in this paper. All-pairs comparisons are used on evals across several datasets differing in temperature, random seed, and evals/models included. The resulting \texttt{Plotly} figures interactively show more details like the model pair and various noise types. 
The rest of this section explain the data visualization tool and what it shows with a single example. If in doubt, the reader should sample a few evals and verify that our observations hold.
See \S\ref{sub:exceptions} for a discussion of some exceptions.

\subsection{Exploratory analysis tools and findings}
\label{sec:exploratorytools}
\subsubsection{Data heatmap}
\label{sub:heatmap}
\begin{figure}[t]
    \centering
    \includegraphics[width=0.49\linewidth, trim={80pt 165pt 40pt 0pt}, clip]{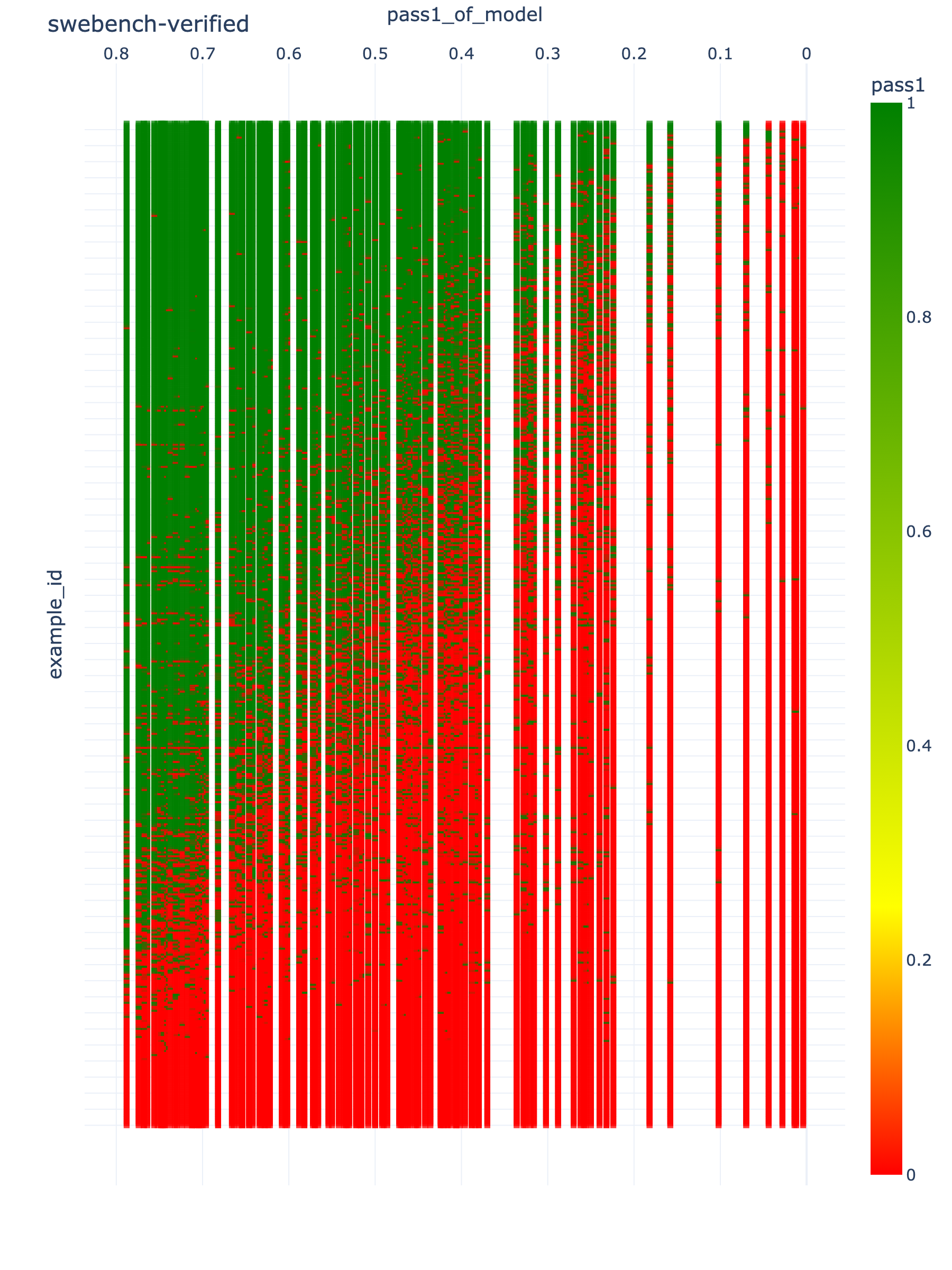}
    \includegraphics[width=0.49\linewidth, , trim={80pt 165pt 40pt 0pt}, clip]{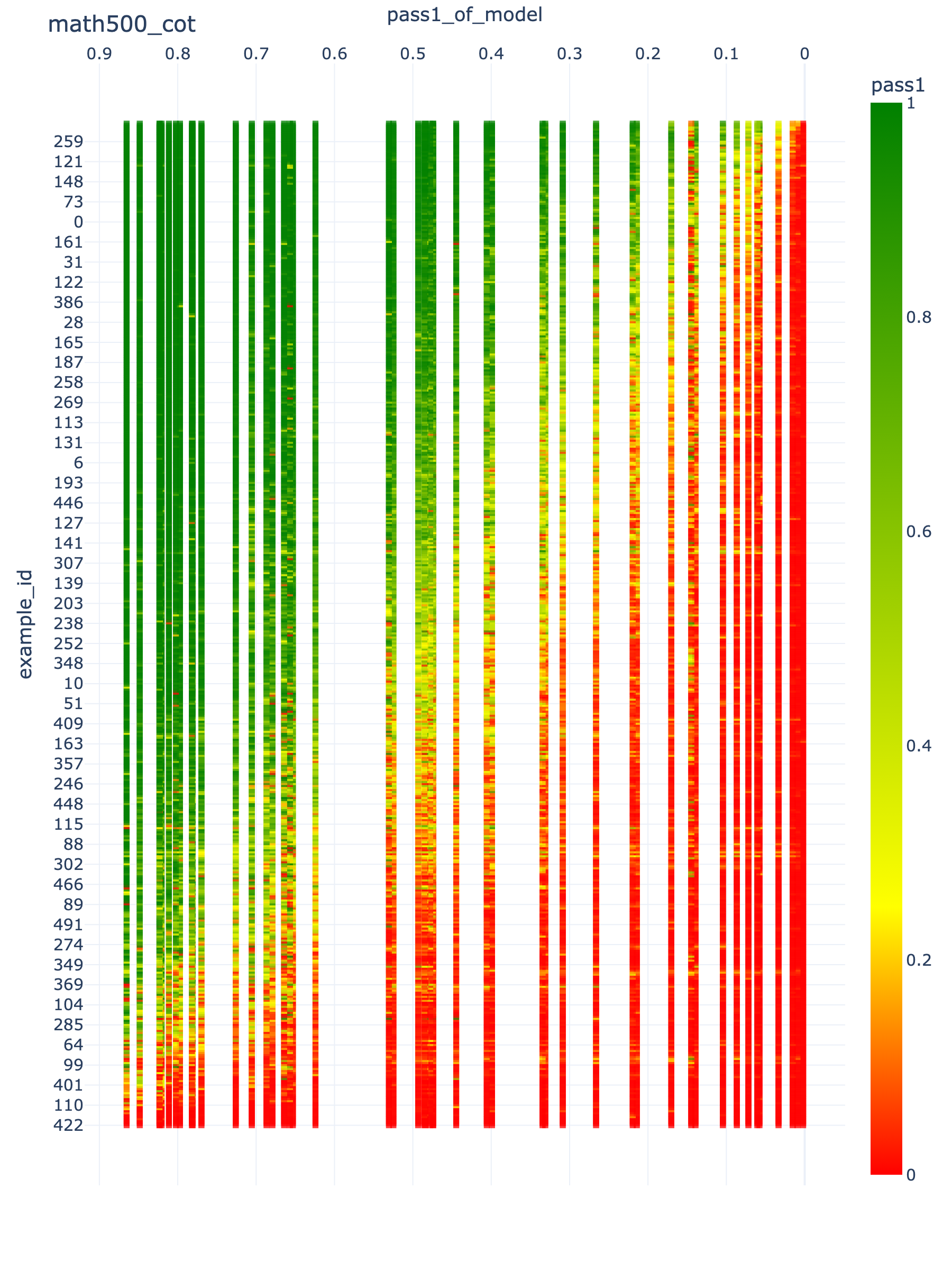}
    \caption{Results heatmap. Each row is a different question, sorted from the easiest to the hardest. Each column is a model whose $x$-coordinate is the overall accuracy. Left: SWEbench-verified, 1 prediction per question; Right: MATH500, 1000 predictions per question. See \S\ref{sec:exploratorytools} for more details.}
    \label{fig:heatmap}
\end{figure}

To get an overview of our data, we use a heatmap shown in Figure~\ref{fig:heatmap}. Each row is a different question, sorted from the easiest to the hardest. Each column is a model whose $x$-coordinate is the overall accuracy $\E[A]$ of this model. Like models, we might consider showing questions with $y$-coordinate at their average accuracy instead of rank, but that would depend on which models are included, and less intrinsic to the eval. We immediately observe that models at a similar overall accuracy also do similarly on individual questions. 
Around 1000 samples are drawn for each question and model for MATH500 in Figure~\ref{fig:heatmap}. The SWEbench figure is based on the leaderboard data, which had to commit to single answers, but the same overall pattern holds, though with more noise. On MATH500, most of the red regions of bad models are still non-zero.

\subsubsection{Noises from all-pairs comparison}
\label{sub:allpair}

Figure~\ref{fig:pairedSE} shows that the total SE agrees well with the $\operatorname{Beta}(p, 1-p)$ theory predictions described in \ref{sub:beta} and the range of the data noise and prediction noises. Only model pairs that are close in performance are plotted $|\E[A]-\E[B]| < 5\SE(A-B)$ to avoid pointless comparisons between models whose difference is not in doubt. 

\begin{figure}[th]
    \centering
    \includegraphics[width=0.49\linewidth, trim={5pt 50pt 30pt 50pt}, clip]{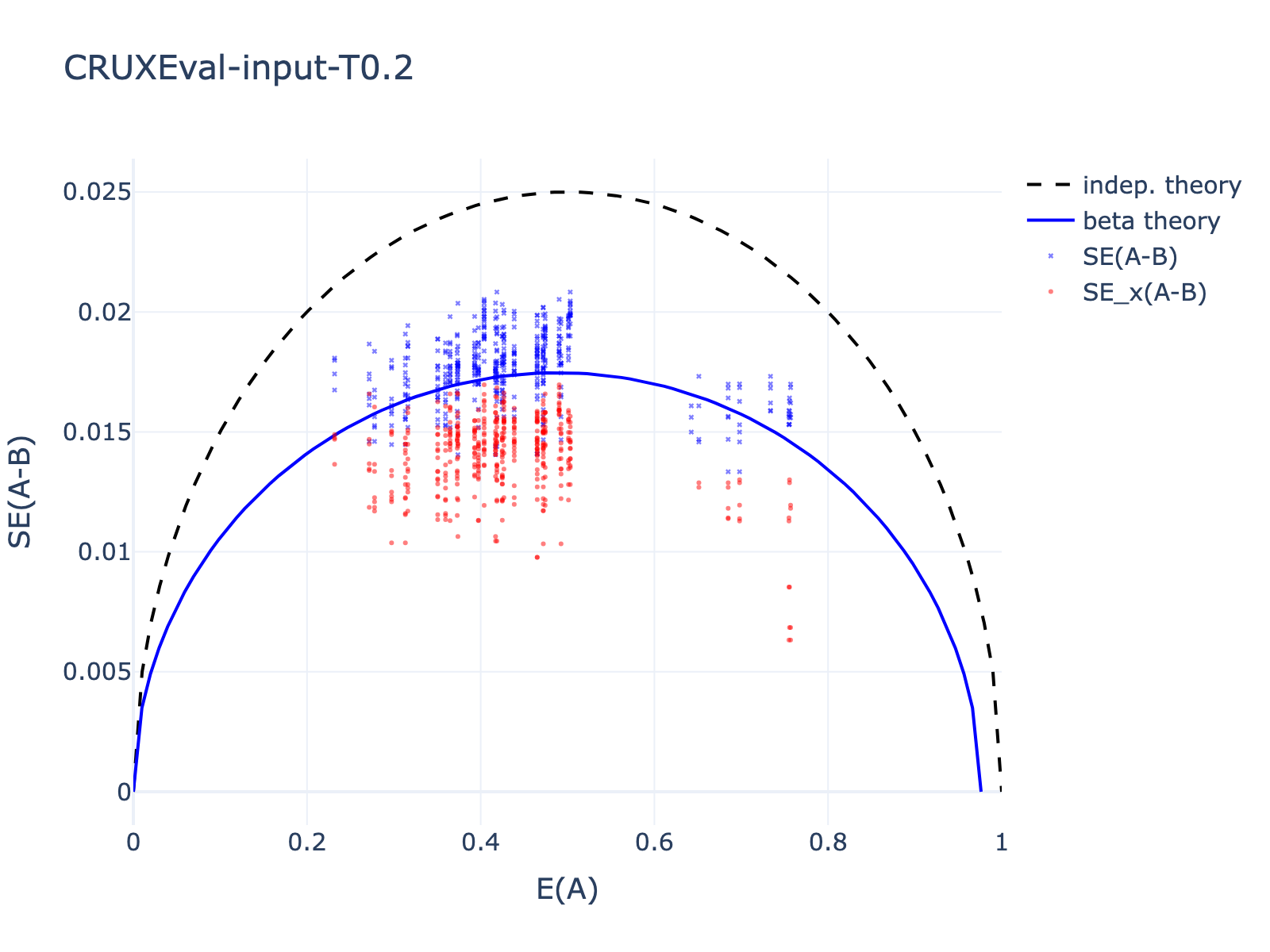}
    \includegraphics[width=0.49\linewidth, trim={5pt 50pt 30pt 50pt}, clip]{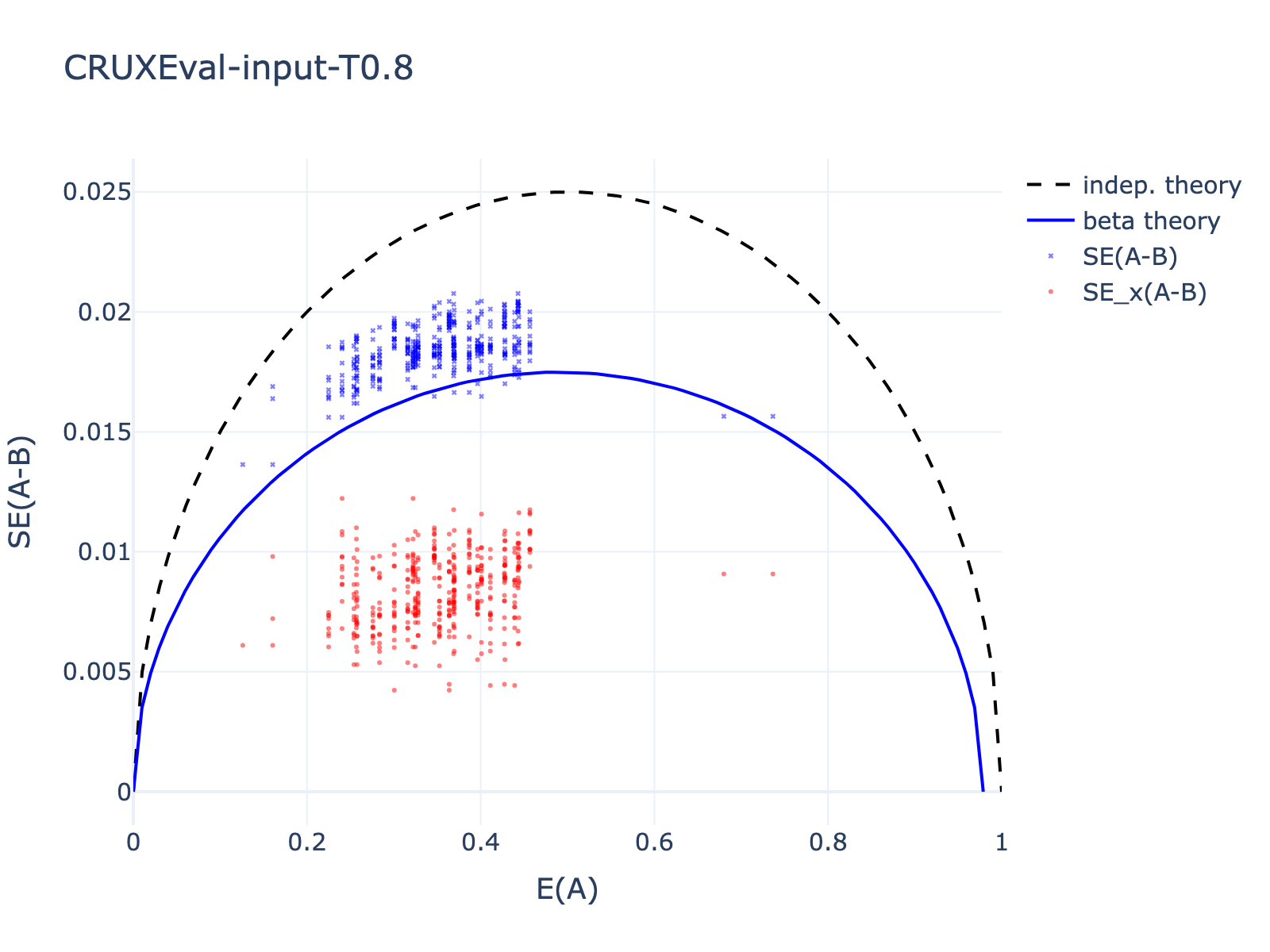}
    \vspace{-10pt}
    \caption{All noise components on CRUXEval at temperature 0.2 (left) and 0.8 (right). The prediction noise dominates at 0.8 temperature, whereas the data noise dominates at 0.2 temperature, while both still have similar total noise.}
    \label{fig:allnoise}
\end{figure}

The noise components can be traded off to some extent so they can depend on the setting and not as predictable as the total. We find at more natural temperatures of $[0.7, 1]$, the prediction noise tends to be higher than the data noise across evals. Figure~\ref{fig:allnoise} shows the noise components at two different temperatures on CRUXEval, where the breakdown differs but the total noise remains the same.

\subsubsection{The Beta theory makes good noise predictions and fits the data}
\label{sub:beta}
The $\Beta(p, 1-p)$ distribution, where $p$ is the model's mean accuracy, fits the empirical distribution of question-level expected accuracies well and predicts a paired prediction variance of $p(1-p)$, explaining the observed trend in Figure~\ref{fig:pairedSE}. The details are in \S\ref{sec:A.beta}.

\subsubsection{SE vs. difference}
\label{sec:se_vs_diff}
\begin{figure}[t]
    \centering
    \includegraphics[scale=0.13, trim={0pt 50pt 322pt 40pt}, clip]{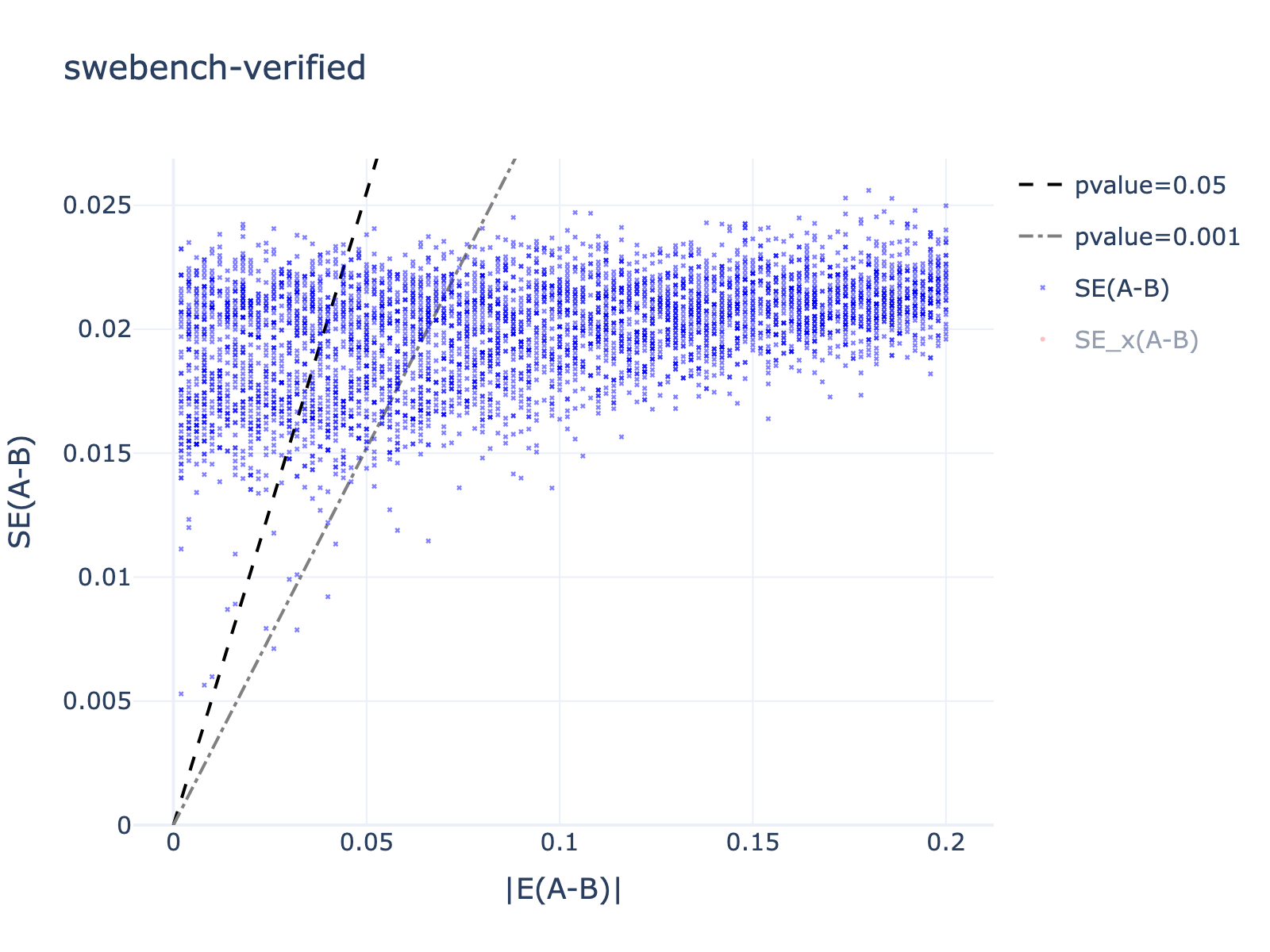}
    \includegraphics[scale=0.13, trim={0pt 50pt 26pt 40pt}, clip]{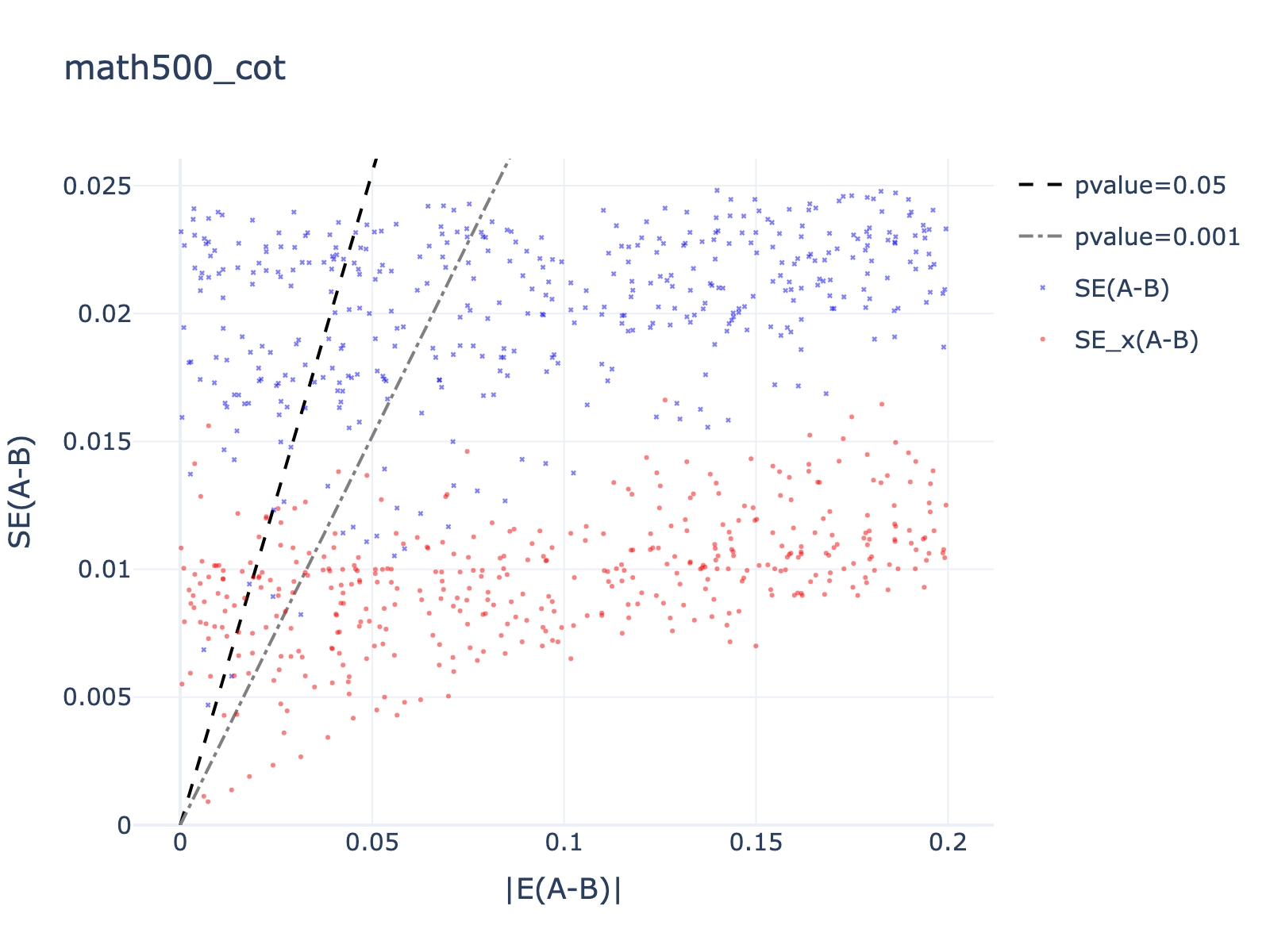}
    \caption{\textsf{SE(A-B)} vs. the difference \textsf{E(A-B)}. constant p-values are straight-lines on this figure and every model pair to the right of the line are more significant.
    }
    \label{fig:se_vs_diff}
    \vspace{-10pt}
\end{figure}

Figure~\ref{fig:se_vs_diff} plots the paired standard error $\SE(A-B)$ vs. the difference $\E(A-B)$ to show the range where the stat. sig. is questionable and exactly which pairs are significant.

\begin{example}[Significance of training seeds]
The variance due to training seeds is an intuitive standard, but does not correspond to the statistical standard. In theory, some training seeds can cause training to fail completely and produce a very significant result. Empirically this indeed happens where the training seed can produce more significant results (Figure~\ref{fig:se_vs_diff_seed}). We can check which situation we are in with this figure instead of intuitively adopting the variation of training seed \cite{madaan2024variance, heineman2025signalnoise} as the standard.

\begin{figure}[th!]
    \centering
    \includegraphics[scale=0.13, trim={0pt 50pt 322pt 40pt}, clip]{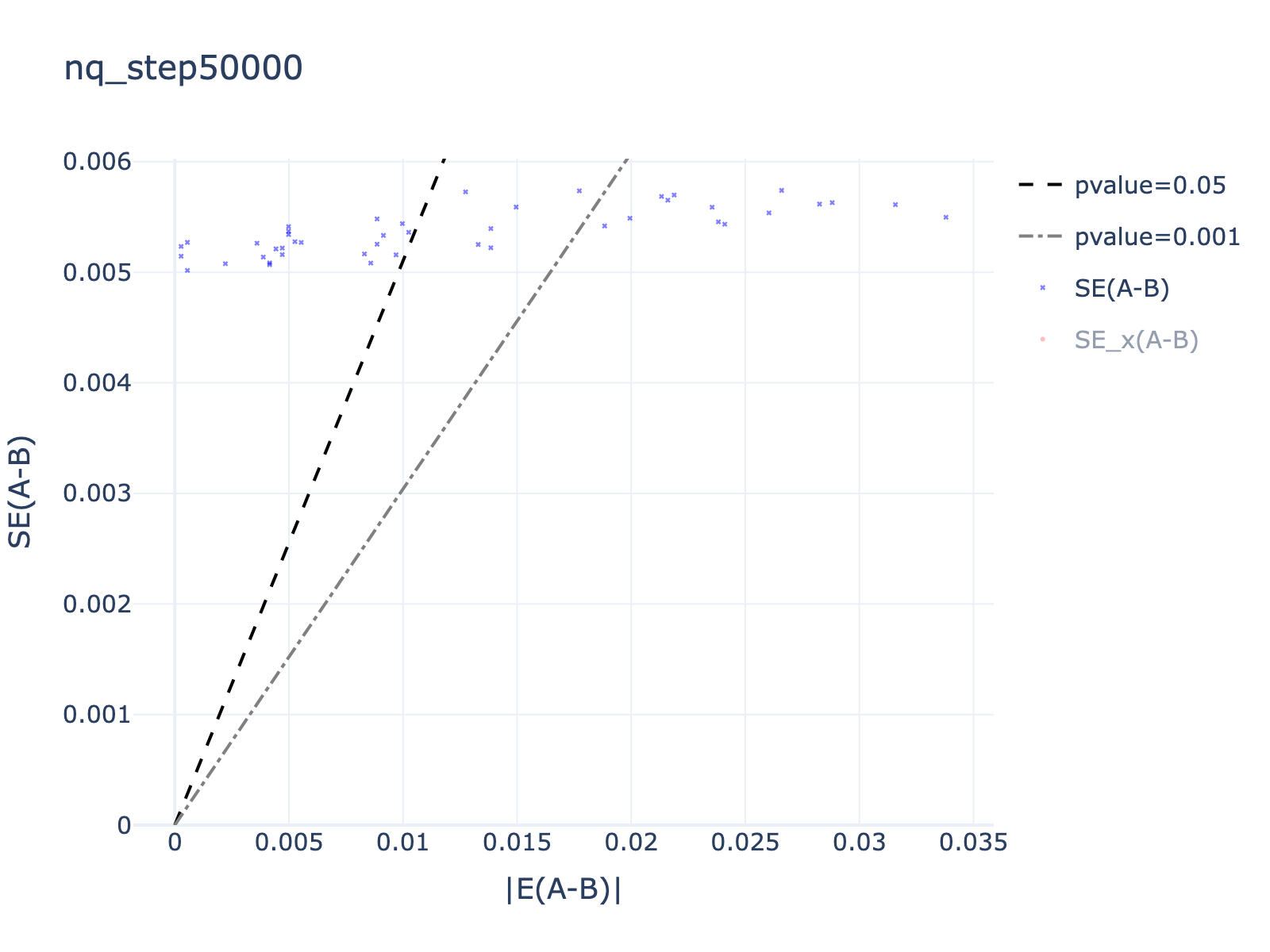}
    \includegraphics[scale=0.13, trim={0pt 50pt 26pt 40pt}, clip]{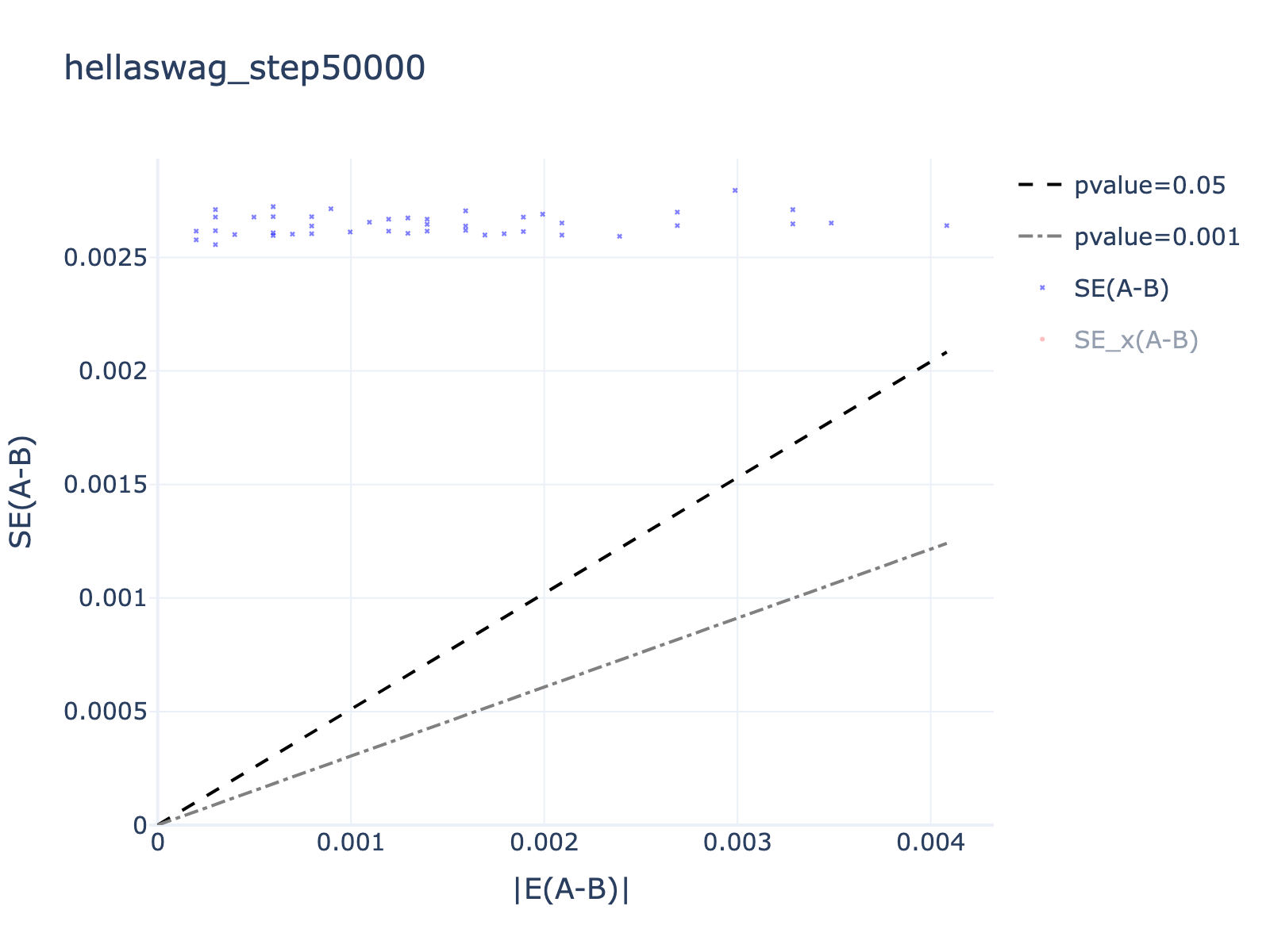}
    \caption{\textsf{SE(A-B)} vs. the difference \textsf{E(A-B)}. Each point is a pair of random training seeds. Under the null hypothesis, the proportion of points on the right side of each $p$-value line is expected to be $p$. The eval nq produced more significant results than the null hypothesis whereas hellaswag produced even less than the null. constant p-values are straight-lines on this figure and every model pair to the right of the line are more significant.  
    }
    \label{fig:se_vs_diff_seed}
    \vspace{-10pt}
\end{figure}

\end{example}

%% file: discussions.tex
\section{Discussions}
\label{sec:discussions}

\subsection{Applications and Recommendations}
\label{sub:recs}
In addition to the recommendation of \citet{miller2024error}, we advocate for more global noise measurements to decouple noise analysis from most experiments using all-pairs comparisons, so practitioners can focus on their experiments instead of the noise analysis.
For some motivating examples, in the Llama 3 report many tables show the unpaired standard errors, which is too loose and gives no additional information since they are just a simple function of the accuracy. \citet{merrill2026terminalbench} used the prediction noise for their confidence intervals (incorrect) and \citet{chan2025mlebenchevaluatingmachinelearning} measured the total noise with different number of seeds (correct missing context).
We recommend that eval and leaderboard builders run the all-pairs comparison to measure all the noise types on their eval, thus providing well-specified noise measurements and taking responsibility for the statistical reliability of their evals. This can even be done by a third-party as in this work, as long as leaderboard maintainers release their question level results like many leading evals \cite{jimenez2024swebench, jain2024livecodebench, merrill2026terminalbench}, ideally with multiple predictions per question so the noise components can be estimated.
This might be a unique possibility in the LLM paradigm, where both models and evals have global identities to enable this approach.

For experimeters, we recommend in increasing accuracy and overhead:
\begin{itemize}[leftmargin=*,topsep=-0.25em,itemsep=-0.3em]
    \item use the heuristics $\SE[A-B] \approx \SE[A]$ and read these SEs from our main table
    \item read our paired total noise curves (Figure~\ref{fig:pairedSE}) to obtain the ranges of each type of noise. Total noise levels can be used when evaluating using one prediction per question and data noise can be used to determine how effective averaging is.
    \item use method and estimators to measure the noise components on any data, which enables the detection of much smaller signal in controlled experiments.
\end{itemize}

If we see that prediction noise is dominant, then averaging can be used to greatly increase the power, and the usual method of inferring noise from the training curve can also be justified.

Finally, with these measurements, we can have noise-aware meta-analysis of multiple evals \S\ref{sec:metaanalysis}.

\subsection{Solving hard and special problems}
\label{sub:hardproblems}
Leading agentic evals often have fewer than 1000 questions, with terminal bench \cite{merrill2026terminalbench} having fewer than 100 questions. In contrast, 100 multiple choice questions clearly have a lot of noise when questions are guessed.
Perhaps the reason for believing that very small evals are okay in the agentic setting is that the setting can capture very hard problems that are not guessable. For example, if a model solved an important open problem then we would not object to the sample size of one. Here the prior probability of success is near 0, so while observing a success gives a lot of information, we are expected to fail and get no information. In other words, either we get no information or the problems are not that special.
More directly, one can use the mixed predictor \eqref{eq:mixpred} to generate variance from the differences in model predictions once there are some successes, just like guessing. Such evals still need to satisfy the regular statistical standard to be meaningful after any initial surprise. 

In human evaluations, we seem to get a lot of signal by evaluating on hard problems with a number of steps (e.g. interviews, math). The key is to reflect on the details of how the problem was solved, and if it seems to be memorized. In contrast, LLM evals and RLVR settings only give 1 bit back after doing a lot of work. If we continue in the direction of smaller and harder evals, we probably need to also generate more information per question.

\subsection{Caveats of the expected metric}
Averaging is statistically valid and can be recommended to get more power on controlled experiments. However, this is an unusual standard not available to human testing and scientific experiments. \S\ref{sec:caveats_exceptions} discusses these caveats especially when targeting a particular eval with aggregation method or when the variable affects the distribution's sharpness.

\subsection{Empirical findings}
\label{sub:mainremarks}
\paragraph{Predictable total noise.}
When the prediction noise is much greater than the data noise, then total noise $\approx$ prediction noise, and prediction noise is predicted by the Beta theory, so the predictable total noise is a consequence of prediction noise $\gg$ data noise.
However, Figure~\ref{fig:allnoise} shows non-negligible data noise, yet the total noise still follows the Beta theory prediction even when the data noise exceeds the prediction noise at low temperature, except for chain-of-thought predictions.
More data is needed to further test if this continues to hold, and if low temperature can reduce prediction noise as the model is training.

\paragraph{Prediction noise $>$ data noise.} It is not surprising that current LLMs are similar to each other on question level metrics, as they are trained on similar data using similar architecture and learning methods. So it is expected that LLMs perform similarly on each question if they have similar overall abilities. In theory, the prediction noise can still be arbitrarily small, but LLMs seem to be both positively correlated and have sufficient prediction noise for this result to hold. 

Large prediction noise has impact on modeling question difficulty using the Item Response Theory (IRT) or filtering questions \cite{polo2024tinybenchmarks}. Section 5 of \citet{madaan2024variance} and \ifarxivversion\citet{evalarena}\fi{} find that IRT does not work well. The data filtering method $\tau-$ of \ifarxivversion\citet{evalarena}\fi{} finds questions that are inversely correlated with the overall model performance.
When there is high prediction noise, there is no way to get a reliable signal from a small number of questions or to reliably blame the questions for noise.

\subsection{Limitations}
While the method is general, all the empirical data are from correctness evals, so how well the empirical findings generalize to other evals remains to be tested.
Currently, we did not consider the potentially impactful clustered correction of \citet{miller2024error}. 

\ifarxivversion
\paragraph*{Acknowledgments.}
I thank Sean O'Brien, Lovish Madaan, Dieuwke Hupkes, Weina Xie, Alex Gu, Jiawei Liu, Yuhang Lai, Linyuan Gong, and Sten Sootla for making question-level data available for analysis. I am grateful to Evan Miller, Ziang Xie, Nicolas Usunier, Zach Rait, Yuxiang Wei, Jannik Kossen, and Ari Holtzman for valuable discussions and feedback; Pedro Rodriguez, Ofir Press, Naman Jain, Baptiste Rozière, Gabriel Synnaeve, Dawn Song, and Zijian Wang for their advice and support. The start of this project is inspired by Chatbot Arena and the clarity of \citet{miller2024error} greatly helped in finishing it.

\subsection*{AI use statement}
In this work, we used generative AI tools for coding, debugging, checking the math, formatting references. It was very helpful in debugging a tricky issue to get an unbiased estimator to pass the total variance test. An initial draft of of the estimator testing appendix section was generated by AI from the code.
We have reviewed all AI-assisted work. We take responsibility for the final content of this work,
including text, claims or artifacts produced with the aid of generative AI.
\fi

%% file: A-method.tex
\section{Method details}
\label{sec:A.method}

\subsection{Estimating the prediction noise directly vs. using quesiton-level results}
\label{sub:alternative_decompose}
For a infinite set of questions, presumably we have no noise where $\var_\seed[\E_x[A]]=0$, so we consider $N$ specific questions with $\E_x [\var_\seed[A(x, \seed)]$ being the prediction noise that we can measure directly
\begin{align}
\var_\seed[\E_x[A]] = \var_\seed \left[\frac1{N}\sum_{i=1}^N A(x_i, \seed)\right] = \frac1{N^2} \sum_{i=1}^N \var_\seed[A(x_i, \seed_i)] = \frac1{N} \E_x [\var_\seed[A(x, \seed)]. \label{eq:prednoise}
\end{align}
Here $\E_x[A]$ is taken over the uniform distribution of a finite $N$. While the two sides are equal, estimating the LHS directly with a finite number of random seeds $\seed$ can have a much higher error. The estimator corresponding to the LHS is running the eval $K$ times and measure the variance of the $K$ final results $\operatorname{var}(\bar{A}_1, \ldots, \bar{A}_K)$ where $\bar{A}_j = \frac1{N}\sum_{i=1}^N A(x_i, \seed_{ji})$. This worse estimator fails to use the fact that random samples are draw on each question independently. 
The following  example shows this concretely, where the variance of the LHS is about 500 times bigger than the RHS.

\begin{Verbatim}
Let X be a 100x5 matrix of Bernoulli(0.5) random variables.
Draw 10000 samples of X to calculate
LHS = var(mean(X, dim=0), dim=1),
RHS = 1/N * mean(var(X, dim=1, ddof=1), dim=0)
Show the mean, variance and the first 5 elements of the two lists
\end{Verbatim}

Conversely, the middle equality holds only if $\seed$ acts independently on the questions so that $A(x_1, \seed)$ is independent of $A(x_2, \seed)$ as if they use separate seeds. This holds for prediction seeds but not for model training seeds. For random seed that produce correlation, the RHS is still the variance of resampling of the seed for each question, but is not equal to the LHS.



\subsection{Estimating from samples}
\label{sec:sample_estimator}
While \eqref{eq:paired_total} allows us to estimate the variance components directly from $K$ paired samples $A(x, \seed_j) - B(x, \seed_j)$ for $j = 1, \ldots, K$, this is inaccurate since the seeds or predictions are interchangeable.
We can use this to derive formulas for estimating from samples accurately, as if using all $K \times K'$ pairs of collected predictions $A(x, \seed_j) - B(x, \seed'_k)$ for $j = 1, \ldots, K$ and $k \in 1, \ldots, K'$. First, the prediction variance $\E_{x}[\var_\seed[A - B]]$ decomposes by independence on each question $x$,
\begin{align}
\var_{\seed}[A(x, \seed) - B(x, \seed')] &= \var_{\seed}[A(x, \seed)] + \var_\seed[B(x, \seed)],\\
\E_x[\var_{\seed}[A(x, \seed) - B(x, \seed')]] &= \E_x[\var_{\seed}[A(x, \seed)]] + \E_x[\var_\seed[B(x, \seed)]].
\end{align}

Next, the data variance $\var_{x}[\E_\seed[A - B]] = \var_{x}[\E_\seed[A] - \E_\seed[B]]$ decomposes by linearity of expectation.
Finally, the total variance is slightly tricky, where
\begin{align}
\var_{x, \seed}[A - B] = \var_{x, \seed}[A] + \var_{x, \seed}[B] - 2 \cov_{x, \seed}[A, B] \\
= \var_{x, \seed}[A] + \var_{x, \seed}[B] - 2 \cov_{x}[\E_\seed[A], \E_\seed[B]]. \label{eq:paired}
\end{align}
The equality is by the law of total covariance, $\cov_{x, \seed}[A, B] = \cov_{x}[\E_\seed[A], \E_\seed[B]] + \E_x[\cov_{\seed}[A, B]]$ where the second term is 0 since different random predictions on a given $x$ are independent.

\begin{figure}[hbt]
    \centering
    \includegraphics[width=0.9\linewidth, trim={5pt 15pt 5pt 5pt}, clip]{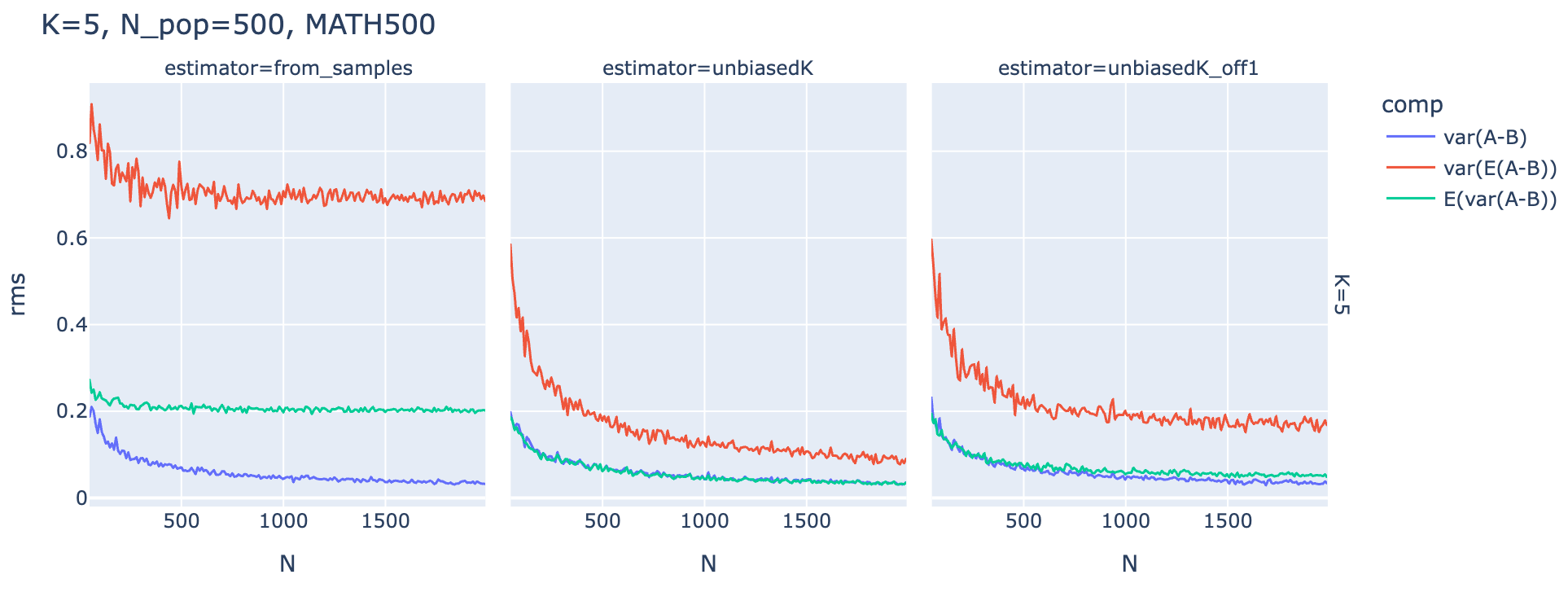}
    \vspace{-0.5em}
    \caption{Relative errors of the paired variances.  $K=5$ samples are drawn from the 500 questions from MATH500, where 1000 real predictions are drawn from models $A$ and $B$ on each question, which is treated as the ground truth. The root mean squared relative errors (rms) of variance components are plotted vs. bootstrap sample size $N$, extending beyond the population size 500. The left figure without the correction on $K$ has an unacceptably high 70\% relative error even with 2000 data points. The middle figure is with the correction of \S\ref{sec:sample_estimator} and the right figure uses $1/K$ instead of $1/(K-1)$.}
    \label{fig:acc_estimator}
\end{figure}

\paragraph{Small $K$ correction.}
To estimate $\var_x[\E_\seed[A]]$ from $K$ sampled predictions per question, a correction from the direct estimator can significantly increase the accuracy. We describe the concept for the unpaired case for simplicity but it matters the most in the paired case. By the law of total variance on $A^{(K)}(x) = \frac1{K} \sum_{j=1}^{K}A(x, \seed_j)$, we have
\newcommand{\Abar}{A^{(K)}}
\begin{align}
\var_{x, \seed}[\Abar] &= \var_x[\E_\seed[\Abar]] + \E_x[\var_\seed[\Abar]] \\
&= \var_x[\E_\seed [A]] + \frac1{K} \E_x[\var_\seed[A]] \\
\implies \var_x[\E_\seed [A]] &= \var_x[\Abar]  - \frac1{K} \E_x[\var_\seed[A]] \\
&\approx \hat{\var}_x[\Abar] - \frac1{K-1}\hat{\E}_x\hat{\var}_\seed[A]
\end{align}
where the hats mean sample estimate. This correction for small $K$ is important because we will average over $N$ questions. If the estimate on each question is biased by $1/(K-1)$, the average error will still be off by the bias no matter how big $N$ is. Whereas if the estimate is unbiased on each question, then the average error can decrease as $N$ increases. Figure~\ref{fig:acc_estimator} shows this observation on MATH500, where the small-$K$ correction is needed to reduce the error to an acceptable level even with $2000$ bootstrapped questions. The high relative error is because the paired data noise is much smaller than the prediction noise.
An instructive example is to consider when $\E_\seed[A-B]=0$ but an estimate using $K$ samples will not be 0 due to non-zero prediction noise. 

\paragraph{Same model correction.}
Using \eqref{eq:paired} to compute $\var_{x,\seed}[A(x, \seed_1) - A(x, \seed_2)]$ is too small when using the same set of samples for $A$. Instead, we can use the following equality and an unbiased estimator for $\E_{x}[\var_\seed[A]]$ with $A = A(x, \seed_1)$, $B = A(x, \seed_2)$.
\begin{align}
\var_{x, \seed}[A - B] &= \var_{x}[\E_\seed[A - B]]
+ \E_{x}[\var_\seed[A - B]] \\
 &= 0
+ \E_{x}[\var_\seed[A] + \var_\seed[B]] \\
&= 2 \E_{x}[\var_\seed[A]]
\end{align}

Equivalently, we can still use \eqref{eq:paired} and Table~\ref{tab:paired} if we sample without replacement when estimating $\E[A(x, \seed_1)A(x, \seed_2)]$ to avoid the bias over-estimating the covariance using the same set of samples.
Suppose we have $K$ samples for a question $x$ with scores $A_1, \ldots, A_k$. To estimate $\E[A(x, \seed_1)A(x, \seed_2)]$ we must use $\texttt{mean}_{i,j \neq i} A_i A_j = \frac1{K^2 - K} \sum_{i, j \neq i} A_i A_j = \frac1{K^2 - K} [(\sum_i A_i)^2 - \sum_i A_i^2]$.

\subsection{Implementation in array notation}
\label{sec:implementation}
In an experiment, we typically evaluate the model on all $N$ questions, and may draw $K$ answers for each question. In this section, we overload our notation with actual samples $A, B \in \R^{N \times K}$ and present the formulas in \S\ref{sec:sample_estimator} in numpy-style code. For simplicity, we use the same number of predictions $K_i = K$ on all questions $x_i$. For the unpaired case, we have Table~\ref{tab:unpaired}. For paired, we have Table~\ref{tab:paired}.

\begin{table}[h]
    \centering
    \begin{tabular}{|l|l|l|l|}
        \hline
        \textbf{Name} & \textbf{Formula} & \textbf{Code} & \textbf{Bernoulli} \\
        \hline
        total variance & $\var_{x, \seed}[A]$ & $\texttt{var(A)}$ & $\bar{p}(1-\bar{p})$  \\
        data variance & $\var_x[\E_\seed[A]]$ & $\texttt{var(mean(A, axis=1))-b}$ & $\frac1{N}\sum_i (p_i - \bar{p})^2 $  \\
        prediction variance & $\E_x[\var_\seed[A]]$ & $\texttt{mean(var(A, axis=1))+b}$ & $\frac1{N}\sum_i p_i(1-p_i)$  \\
        \hline
    \end{tabular}
    
    where $\texttt{b=1/(K-1)*mean(var(A, axis=1))}$
    \caption{Unpaired estimators where $A \in \R^{N \times K}$ with $K$ samples for each of $N$ questions. }
    \label{tab:unpaired}
    \begin{tabular}{|l|l|}
        \hline
        \textbf{Formula} & \textbf{Code}  \\
        \hline
         $\var_{x, \seed}[A - B]$ & \texttt{var(A)+var(B)-2*cov1(mean(A,axis=1),mean(B,axis=1))} \\
        $\var_{x}[\E_\seed[A - B]]$ & \texttt{var(mean(A,axis=1)-mean(B,axis=1))-b}\\
         $\E_{x}[\var_\seed[A - B]]$ & \texttt{mean(var(A,axis=1)+var(B,axis=1)) +b} \\
    \hline
    \end{tabular}
    
    where $\texttt{b=1/(kA-1)*mean(var(A, axis=1))+1/(kB-1)*mean(var(B, axis=1))}$
    \caption{Paired estimators where $A \in \R^{N \times K_A}$, $B \in \R^{N \times K_B}$. }
    \label{tab:paired}
\end{table}




%% file: A-beta.tex
\section{The Beta theory predicts the noise and fits the distribution}
\label{sec:A.beta}

\begin{figure}[h]
    \centering
    \includegraphics[width=0.49\linewidth, trim={35pt 35pt 15pt 35pt}, clip]{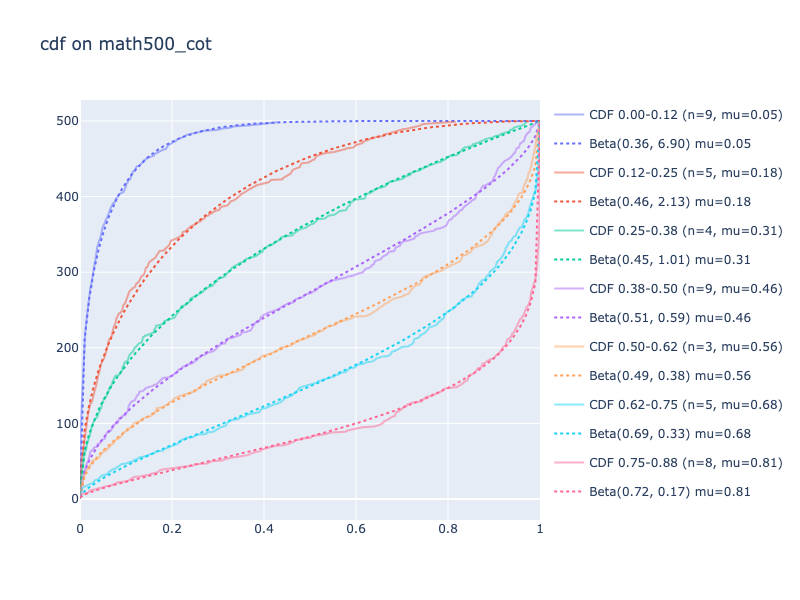}
    \includegraphics[width=0.49\linewidth, trim={35pt 35pt 15pt 35pt}, clip]{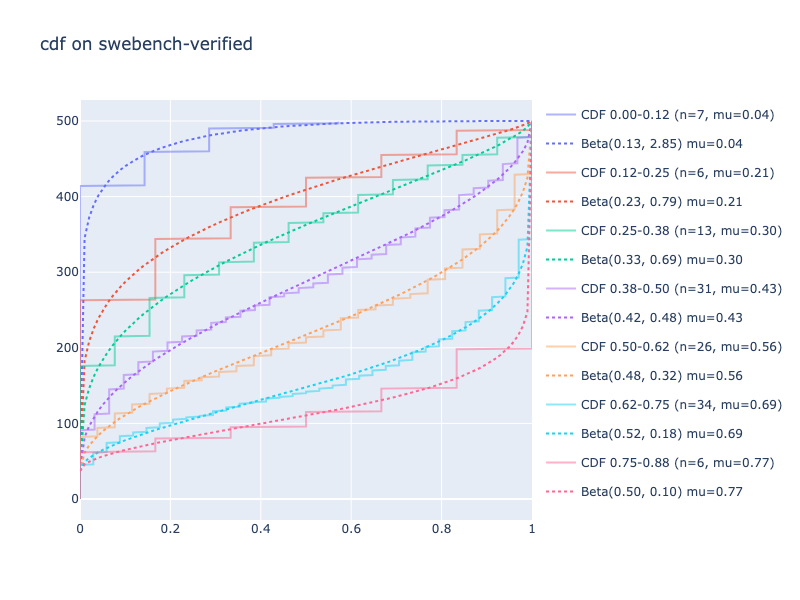}
    \caption{Empirical cumulative distribution curves (CDF, solid) and their respective Beta models (dots). Each color represents a different bin by model accuracy. The average accuracy of each question in each bin of models is computed and sorted to produce the CDF curves. The fit is good.}
    \label{fig:betacdf}
\end{figure}

Figure~\ref{fig:betacdf} shows the comparison between a $\Beta(a, b)$ distribution and the empirical cumulative distribution (CDF) of the metrics, where the models are binned by range of accuracy. Then $\hat{p}_1 = c_1/K, \ldots, \hat{p}_N=c_N/K$ is computed for each question $i$ across the $K$ combined predictions among the binned models, $c_i$ is the total number of predictions that evaluated as correct on question $i$. Smaller $a+b$ means more bimodal where $a+b$ tend to decrease when the model accuracy is higher, showing that better models become more bimodal whereas very bad models is usually unimodal. For an extreme example, guessing uniformly on multiple choice questions with $C$ choices means the expected accuracy $u_i = 1/C$ for all questions, which has a single mode at $1/C$.

The Beta theoretical model says that the expected accuracy of each question follows the $\Beta(p, 1-p)$ distribution where $p$ is the mean of the model. More precisely, the expected accuracy of question $x_i$ is $u_i \sim \Beta(p, 1-p)$ so that $A(x_i), B(x_i) \sim \operatorname{Bernoulli}(u_i)$ and $\E[A(x_i)] = u_i$. To see that this leads to the observed SE, the models draw independent predictions on each $x_i$ so $\cov_\seed[A,B \mid x_i] = 0$. Thus we have $\E_x[\Var_\seed[A - B]] = \E_x [2\var_\seed[A]]$, which integrates to
\begin{align}
\int_0^1 2u(1-u) \frac{u^{p-1} (1-u)^{(1-p)-1}} {B(p, 1-p)} du = 2 \frac{B(p+1, (1-p)+1)}{ B(p, (1-p))} = p(1-p),
\end{align}
where $B(a,b) = \int_0^ 1 u^{a-1}(1-u)^{b-1} du$ is the normalization constant for the $\Beta(a, b)$ distribution. One consideration to fit the data better is that all models fail on some questions, so we suppose those questions actually have $u_i = 0$ instead of $\Beta(p, 1-p)$. 

%% file: A-caveats.tex
\section{Caveats and exceptions}
\label{sec:caveats_exceptions}
\subsection{Caveats of the expected metric}
\label{sub:standard}
With the impressive increase in power, we recommend averaging for most controlled experiments on model architecture, optimization, loss function, etc. that are not expected to have a large effect on the eval and do not target the eval specifically.
However, there are some caveats if our goal is to improve a particular eval metric using all the best available methods, and when the independent experiment variable is expected to affect the ``sharpness" of the distribution.
Example~\ref{ex:majorityvoting} shows how improvement in expected accuracy leads to no increase after aggregation, and how sharpening the distribution on good-enough model can lead to much better expected accuracy. Example~\ref{ex:verification} shows how increased expected accuracy leads to a decrease the final performance when the trade-off is less diversity, which can happen in post-training. Example~\ref{ex:weak} compares the expected metric vs. alternatives and shows that arbitrarily small but consistent changes is significant using the expected metric, clarifying that an independent standard of effect size is also needed. 

\begin{example}[majority voting]
\label{ex:majorityvoting}
Suppose the expected accuracies on all questions are the same with $\E[A] = 0.6$, $\E[B] = 0.9$ where the expectation averages out the prediction noise. This big improvement in expected accuracy might be trivially achievable by majority voting if we try to put forward the best prediction instead. Whenever the equivalence of sampled predictions can be checked, majority voting can increase the accuracy to 100\% if the equivalence class with the correct answer has the highest probability, which is the case when $\E[A] > 0.5$.
\end{example}

\begin{example}[reliable verification]
\label{ex:verification}
Suppose that on half of the questions $\E[A] = 0.1$ and $\E[B] = 1$, but on the other half of questions $\E[A] = 0.1$ and $\E[B] = 0$. While $\E[B]=0.5 > \E[A]=0.1$ with $B$ having a much higher average, $A$ is better if the predictions can be verified (search problems or formal proofs) where $A$ is 100\% with a verifier but $B$ is only 50\%. If a reliable verifier is available, then the \emph{pass-at-large-K} metric is a more meaningful metric than the average.
\end{example}

\begin{example}[Expected metric and alternatives]
\label{ex:weak}
Suppose we have 100 True/False questions, model $A$ has an expected accuracy of 60\% and Model $B$ 61\% on each question.

\textbf{Expected metric.} We draw enough samples per question to be sure that $A$ is 60\% and that $B$ is 61\%. \textbf{Conclusion}: $A$ and $B$ are significantly different. This metric is used by many papers of the area, and corresponds to the \emph{data noise} after averaging away some prediction noise.

\textbf{Best prediction metric.} Since the probability of correct is greater than half, both models can achieve 100\% accuracy if they predict by majority voting or thresholding. \textbf{Conclusion}: $A$ is the same as $B$. This is used by most classification evals and should be used when pushing the eval.

\textbf{One prediction metric.} One sampled prediction per question is evaluated, so $A$ and $B$ do about 60\% with a $7\%$ standard error on the difference. \textbf{Conclusion}: $A$ is not significantly different from $B$. This metric is based on a single answer per question, is used in most of the sciences and tests for humans, and corresponds to the \emph{total noise}.

Suppose that on question $x_i$, $\E[A(x_i)] = p_i$, $\E[B(x_i)] = p_i + e$, then with enough samples we can conclude that $A$ and $B$ are significantly different no matter how small $e$ is. Still, pure noise without the bias is not expected to produce a significant result.
\end{example}



\subsection{Exceptions to main findings}
\label{sub:exceptions}
We note a few exceptions to the general findings. On guessable evals such as MMLU, when the performance is near random chance, the observed noise is closer to the independent prediction rather than the Beta prediction. 
In Figure~\ref{fig:pairedSE}, swebench-verified, we see 2 outlier results near $x=0.3$ that come from SWE-Fixer~\citep{xie2025swefixer}, due to using a deterministic filter instead of drawing another prediction from the model.
The predictions of Llama on vLLM is more deterministic than expected and thus has higher data noise than others, likely due to faulty settings. Distilled models tend to have less sharp distribution at the same performance level.

%% file: A-recs.tex
\section{Noise-aware meta-analysis of multiple evals}
\label{sec:metaanalysis}
We have an increasing number of evals of varying quality and size, so we want to get more signal from more evals. Directly averaging MMLU where $N\sim10k$ and HumanEval where $N\sim100$ likely leads to worse results than using MMLU alone, so some weighting is necessary.
We advocate using the $z$-score because it is signed and well-behaved.
Recall the $z$-score $z = \frac{\bar{A} - \bar{B}}{\SE[A-B]}$. This work measured the paired and unpaired versions of prediction, data, and total $\SE$s and is suitable here.
Inspecting this list of $z$-scores, one per eval, and checking their consistency is a good start.
To get a meta-$z$-score given the $z$-score of eval $i$, an intuitive method is
$\bar{z} = \frac{\sum_i w_i z_i}{(\sum_i w_i^2)^{1/2}}$,
where some prior beliefs can be incorporated into $w_i$.
For example, $w_i = 1$ means that we believe each eval has equal weight, whereas $w_i = N_i^{1/2}$ means each question has equal weight. While incorporating our prior belief of the importance of each eval into the weight is best suited in LLM evals, $w_i=1$ is recommended as a baseline. See \citet{zaykin2011weightedz} for more related work and analysis, on which we have two remarks specific to LLMs: 1) the effect size of increasing the model size (scaling law) can be a reference on the expected effect size on an eval if desired; 2) giving each question from different evals the same weight is less valid for evals since different evals may capture different abilities, even if this has the most statistical power for a pool of iid questions.

%% file: A-equivalence.tex
\section{Equivalence to the bootstrap and the sign test}
\label{sec:equivalence}
\newcommand{\win}{W}

Two other very well-established methods are the bootstrap \cite{efron1986bootstrap} and paired difference tests such as the sign test \cite{dixon1946sign}. These methods ask if the observed results are likely when the questions are random for bootstrap, and when the comparison outcomes are random for the sign test. We show they all give the same answer.

Bootstrap is a general method where resampling the given questions uniformly with replacement is shown to give the right answer for many estimation problems, including estimating the standard error. That is, \emph{resampling the given questions uniformly with replacement gives the same answer as getting more real questions from the population}. Applied to the problem of noise, the natural question is how likely we are to get the opposite result $\E[A] < \E[B]$ as opposed to the observed result $\E[A] > \E[B]$. If the opposite outcome also has significant probability, then the conclusion is not statistically significant.

The sign test predates bootstrap and instead of randomly resampling the questions, it randomizes the comparison outcomes. It supposes that any difference actually happens with a random probability $\Pr[A > B \mid A(x) \neq B(x)]=0.5$ (null hypothesis) and asks how likely we are to get a more extreme outcome than what we actually observe. The sign test suggests a mix predictor that mixes the predictions of $A$ and $B$ with probability 0.5, which generates variance from different predictions. Let $\hat{y}$ be the prediction to be evaluated then,
\begin{align}
\hat{y}_{AB}(x) = \begin{cases} \hat{y}_A(x) & \text{with prob } 1/2 \\ \hat{y}_B(x) & \text{with prob } 1/2. \end{cases}
\label{eq:mixpred}
\end{align}
If it is quite likely to observe a more extreme outcome under the null hypothesis or equivalently with the mixed predictor, then the conclusion is not statistically significant.

The key step in both methods is estimating the variance of their respective random procedure. Under the same variance, then the probability to observe the opposite result from an observed mean (bootstrap) is also the same as the probability of observing a more extreme result if the real mean is 0 (sign test). The details follow.

To setup both methods, let $\win_A \eqdef \sum_{i=1}^N \1[A(x_i) > B(x_i)]$ be the number of times model $A$ wins against model $B$ and vice versa for $\win_B$, and let's assume $\win_A > \win_B$ for convenience. The question is how likely we have the same conclusion $\win_A > \win_B$ under the randomness prescribed by bootstrap or the sign-test.

\paragraph{The bootstrap.}
We draw another $N$ samples with replacement from the existing samples $x_1, \ldots, x_N$ to obtain $X_A, X_B, X_0 \sim \operatorname{multinomial}(N, q_A, q_B, q_0)$ with $X_A + X_B + X_0 = N$, where the outcome probabilities are $q_A = \win_A / N$ for $A$ win, $q_B = \win_B / N$ for $B$ win, and $q_0 = 1 - q_A - q_B$ for tie. The mean and variance of the resamples are respectively $E[X_A - X_B] = \win_A - \win_B > 0$,
\begin{align}
\text{Var}[X_A - X_B] &= \var[X_A] + \var[X_B] - 2 \cov[X_A, X_B] \\
&= N \left(q_A (1-q_A) + q_B (1-q_B) + 2 q_A q_B\right) \\
&= N \left(q_A + q_B - (q_A - q_B)^2\right) \label{eq:varboots}
\\ 
&\approx N (q_A + q_B) = \win_A + \win_B.
\end{align}
Under bootstrap, a natural question is how likely we would see the opposite result $\Pr[X_A < X_B]$. 
With an approximation that $(q_A-q_B)^2 \ll q_A + q_B$ (else the result is probably significant already), the bootstrap asks if $\Pr[ z \geq \frac{\win_A-\win_B}{\sqrt{\win_A+\win_B}}]$ for standard normal $z$.

Directly estimating the variance from the samples also gives the same answer as \eqref{eq:varboots}. While this is a consequence of bootstrap variance equal to the sample variance, a direct calculation is this
\begin{align}
\var_x[A(x) - B(x)] &= E[(A-B)^2] - E[A-B]^2\\
&= q_A + q_B - (q_A - q_B)^2.
\end{align}


\paragraph{The sign test.}
When comparing model $A$ vs. model $B$, the null-hypothesis is that model $A$ and $B$ each has a 1/2 chance of being better (win) on each question where they differ. $A$ wins if $A$ gets a question correct but $B$ does not, tie if $A$ and $B$ are both correct or incorrect. The question is if the observed results are likely to happen under this null-hypothesis.

The $p$-values is then $\Pr[X \geq \win_A]$ for $X \sim \operatorname{binomial}(\win_A+\win_B, 0.5)$ with $E[X] = \frac{1}{2}(\win_A+\win_B)$ and $\text{Var}[X] = \frac{1}{4} (\win_A+\win_B)$. The question is how likely is $X$ to be as extreme as $\win_A$ as observed. When $\win_A + \win_B$ is large, the normal approximation is accurate and reduces to $\Pr[ z > \frac{\win_A-\win_B}{\sqrt{\win_A+\win_B}}]$ for standard normal $z$ (i.e. one sided). If we also consider how $A$ might be worse, then we should consider the two-sided question. Table~\ref{tab:pairedwinrate} shows our variance components in terms of the win rates $q_A, q_B$ used in the sign test.

So the main empirical variance method, the bootstrap, and the sign test give the same answer. A slight approximation was used on the sign test. Without the approximation, $\var[\text{sign test}] \geq \var[\text{bootstrap}]$ due to using 0.5 for the null hypothesis instead of the empirical win rate. 

\begin{table}[h]
    \centering
    \begin{tabular}{|l|l|}
\hline
\textbf{Formula} & \textbf{Win rates}  \\
\hline
$\var_{x, \seed}[A - B]$ &  $q_A + q_B - (q_A - q_B)^2$ \\
$\var_{x}[\E_\seed[A - B]]$ &  $\frac1{N}\sum_i (q_{A,i} - q_{B,i} - (q_A - q_B))^2 $ \\
$\E_{x}[\var_\seed[A - B]]$ & $\frac1{N}\sum_i q_{A,i} + q_{B,i} - (q_{A,i} - q_{B,i})^2$ \\
    \hline
    \end{tabular}
    \caption{where $A, B \in \{0, 1\}$ with $\Pr[A(x_i) > B(x_i)] = q_{A,i}$, $\Pr[A(x_i) < B(x_i)] = q_{B,i}$. }
    \label{tab:pairedwinrate}
\end{table}



%% file: A-testing.tex
\section{Estimator testing}
\label{sec:estimator_testing}

We validate our variance estimators by sampling from generative models where the ground truth is known. Our primary focus is measuring estimation accuracy and how it scales with sample size, though we also check the bias and test some properties as well.

For each test scenario, we generate data $A, B \in \R^{N \times K}$ from models with known variance components, apply our estimators, and compare against ground truth over many independent attempts. We use two primary metrics: (1) root mean square (rms) relative error to measure accuracy, and (2) $t$-scores to detect systematic bias. 

We test under three generative processes. The \textit{Bernoulli model} samples $N$ problems with replacement from a population of size $N^*$ parameterized by $p_A \in \R^{N^*}$, then sample $K$ Bernoulli trials for each of the $N$ problems. The \textit{stratified Bernoulli model} evaluates when every question is sampled exactly once, simulating the case of drawing samples from real eval, breaking independence and introduces tricky effects. The \textit{bootstrap model} resamples from empirical metric matrices, useful for comparisons to bootstrap and testing based on real data.

\paragraph{Accuracy vs. sample size.}
Our main concern is practical accuracy. Tests establish that estimators achieve rms error below 0.25 for $N = 100$ sample sizes and below 0.13 for larger samples ($N = 400$). We verify this across diverse probability distributions including uniform, Beta, specific distributions and real data. For paired comparisons, tests confirm we get reasonable accuracy (relative rms $< 0.1$) on the total variance and prediction variance. The data variance is the hardest to estimate since it tend to be the smallest where correcting for the bias in $K$ is critical (Figure~\ref{fig:acc_estimator}).

\paragraph{Bias Testing.}
Deriving and testing unbiased estimators proved useful for understanding estimator for each type of sampling procedure. The process of eliminating bias forced us to carefully account for all sources of estimation error—finite sample corrections in both $N$ and $K$, paired sampling effects, and stratified sampling adjustments. Moderately large $N$ is necessary for accurate estimation anyway. We test corrections for $N$ and found they always led to larger errors and thus not recommended. On the other hand, small $K$ is common in practice, so we focus on $K$, and found it to be more than just a bit of bias correction but critical in the case of large $N$ small $K$.

This testing builds confidence that our estimators behave as predicted and that we understand how to get unbiased estimators in each situation, even as we only recommend two of the simplest estimators.